\documentclass[a4paper,fleqn]{cas-dc}

\usepackage[numbers]{natbib}

\def\tsc#1{\csdef{#1}{\textsc{\lowercase{#1}}\xspace}}
\tsc{WGM}
\tsc{QE}
\tsc{EP}
\tsc{PMS}
\tsc{BEC}
\tsc{DE}
\begin{document}
\let\WriteBookmarks\relax
\def\floatpagepagefraction{1}
\def\textpagefraction{.001}
\shorttitle{Maternal Inflammatory Response prediction}
\shortauthors{Sharma et~al.}

\title [mode = title]{Machine learning identification of maternal inflammatory response and histologic choroamnionitis from placental membrane whole slide images}                      
\tnotemark[1]

\tnotetext[1]{Goldstein is supported by National Institutes of Health (NIH) K08EB030120 and R01EB030130. Cooper is supported by NIH U24NS133949, R01LM013523, and U01CA220401. Infrastructure used by this work, including REDCap, is supported by UL1TR001422.
Cooper also reports honoraria from Risk Appraisal Forum, Lynn Sage Breast Cancer Foundation, Jayne Koskinas Ted Giovanis Foundation for Health \& Policy and Consultation for Tempus. The other authors have no relevant financial interest in the products or companies described in this article.}

% \tnotetext[2]{The second title footnote which is a longer text matter
%    to fill through the whole text width and overflow into
%    another line in the footnotes area of the first page.}

\author[1]{Abhishek Sharma}[type=editor,
                        auid=000,bioid=1,
                        prefix=,
                        role=,
                        orcid=https://orcid.org/0000-0001-6666-2179]
\cormark[1]
\ead{as711@northwestern.edu}
% \ead[url]{https://abs711.github.io/}

% \credit{Conceptualization of this study, Methodology, Software}

%\address[1]{, Street 129, 1043 NX Amsterdam, The Netherlands}
\affiliation[1]{organization={Department of Pathology, Northwestern University},
                addressline={750 North Lake Shore Dr.}, 
                city={Chicago},
%               citysep={}, % Uncomment if no comma needed between city and postcode
                postcode={60611}, 
                state={Illinois},
                country={USA}}

\author[2]{Ramin Nateghi}[]

\affiliation[2]{organization={Department of Urology, Northwestern University},
                addressline={675 North St Clair St.}, 
                city={Chicago},
%               citysep={}, % Uncomment if no comma needed between city and postcode
                postcode={60611}, 
                state={Illinois},
                country={USA}}
            
\author[1]{Marina Ayad}[]

\author[1,3]{ Lee A.D. Cooper}[%
   role=,
   suffix=,
   ]
% \ead{wjh@example.org}
% \ead[URL]{https://www.university.org}
\affiliation[3]{organization={CZ Biohub}, 
                city={Chicago}, 
                state={Illinois},
                country={USA}
        }
                
% \credit{Data curation, Writing - Original draft preparation}

\author[1]{Jeffery A. Goldstein}
% \cormark[2]
% \fnmark[1,3]
% \ead{t.rafeeq@example.in}
% \ead[URL]{www.campus.in}

\cortext[cor1]{Corresponding author}

% \cortext[cor2]{Principal corresponding author}
% \fntext[fn1]{This is the first author footnote, but is common to third
% \fntext[fn2]{Another author footnote, this is a very long footnote and
%   it should be a really long footnote. But this footnote is not yet
%   sufficiently long enough to make two lines of footnote text.}

% \nonumnote{This note has no numbers. In this work we demonstrate $a_b$
%   the formation Y\_1 of a new type of polariton on the interface
%   between a cuprous oxide slab and a polystyrene micro-sphere placed

\begin{abstract} 
Introduction: The placenta forms a critical barrier to infection through pregnancy, labor and, delivery.  Inflammatory processes in the placenta have short-term, and long-term consequences for offspring health. Digital pathology and machine learning can play an important role in understanding placental inflammation, and there have been very few investigations into methods for predicting and understanding Maternal Inflammatory Response (MIR). This work intends to investigate the potential of using machine learning to understand MIR based on whole slide images (WSI), and establish early benchmarks. \\ 
Methods: To that end, use Multiple Instance Learning framework with 3 feature extractors: ImageNet-based EfficientNet-v2s, and 2 histopathology foundation models, UNI and Phikon to investigate predictability of MIR stage from histopathology WSIs. We also interpret predictions from these models using the learned attention maps from these models. We also use the MIL framework for predicting white blood cells count (\emph{WBC}) and maximum fever temperature (\emph{$T_{max}$}). \\ 
Results: Attention-based MIL models are able to classify MIR with a balanced accuracy of up to 88.5\% with a Cohen’s Kappa ($\kappa$) of up to 0.772. Furthermore, we found that the pathology foundation models (UNI and Phikon) are both able to achieve higher performance with balanced accuracy of 87.2 \% and 88.5 \% respectively, and $\kappa$ of 0.751 and 0.772 respectively, compared to ImageNet-based feature extractor (EfficientNet-v2s) which achieves a balanced accuracy of 83.7\% and $\kappa$ of 0.724. For \emph{WBC} and \emph{$T_{max}$} prediction, we found mild correlation between actual values and those predicted from histopathology WSIs.\\ 
Discussion: We used MIL framework for predicting MIR stage from WSIs, and compared effectiveness of foundation models as feature extractors, with that of an ImageNet-based model. We further investigated model failure cases and found them to be either edge cases prone to interobserver variability, examples of pathologist's overreach, or mislabeled due to processing errors.
\end{abstract}

\begin{graphicalabstract}
\includegraphics[width=\linewidth]{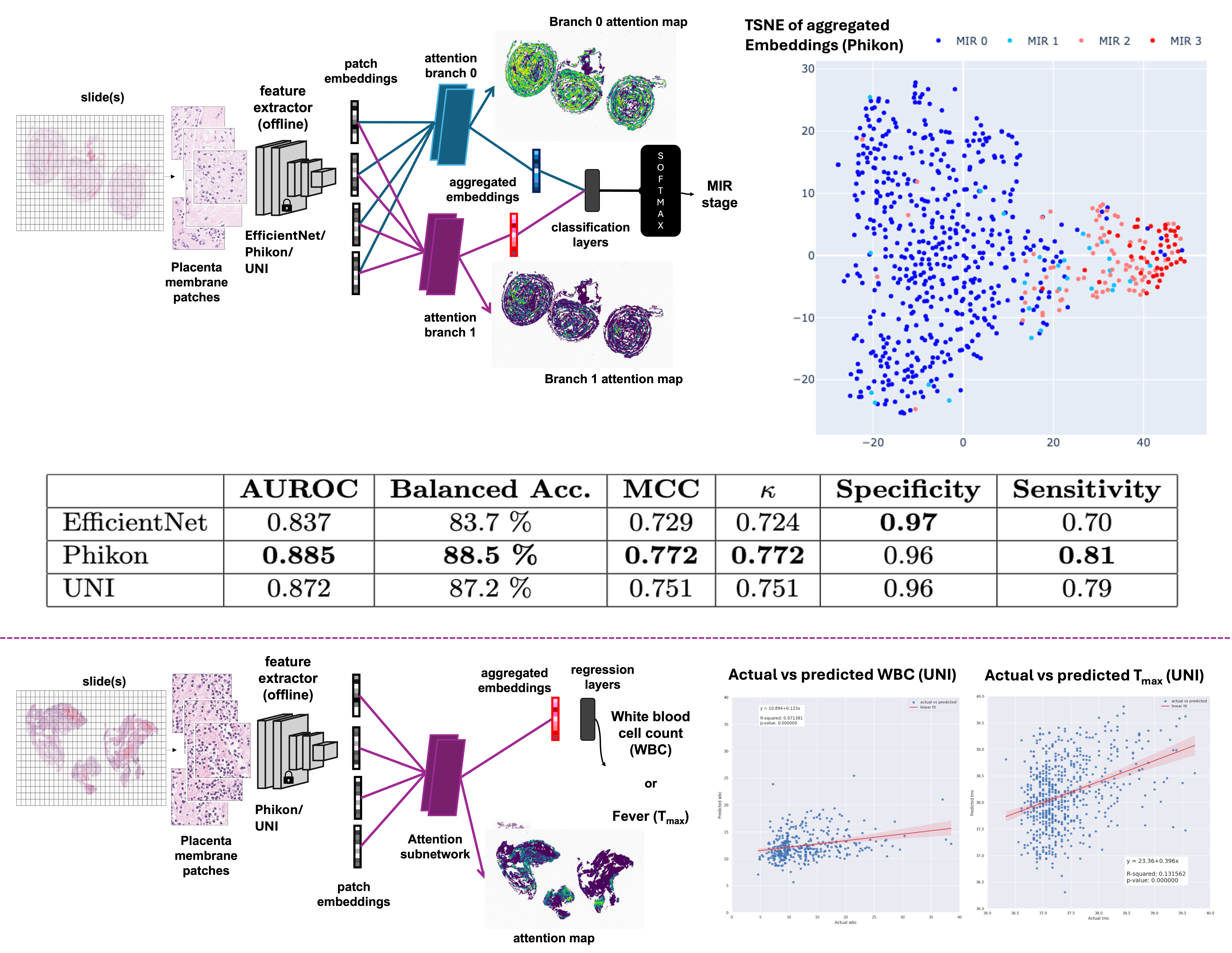}
\end{graphicalabstract}

\begin{highlights}
\item A machine learning model can identify maternal inflammatory response in  placental membranes from whole slide images with high accuracy.
\item Medical imaging foundation models are better feature extractors for MIR identification than model pretrained for ImageNet classification.
\item As an exploratory analysis, we found white blood cell count and maximum fever temperature to be mildly predictable from placental whole slide images.

\end{highlights}

\maketitle
\section{Introduction}
\subsection*{Maternal Inflammatory Response: Definitions and Pathogenesis}
Acute Placental Inflammation (API) is categorized into Maternal Inflammatory Response (MIR) or Fetal Inflammatory Response (FIR) depending on the origin of the inflammatory cells. MIR is characterized by extravasation of maternal neutrophils with movement towards the chorionic layer, followed by moving across the amnion and into amniotic space. MIR is divided into 3 stages. These stages are defined according to the Amsterdam criteria \cite{redline2006inflammatory}. Stage 1 corresponds to subchorionitis, where where maternal neutrophils are limited to the cellular chorion in the membranes and subchorionic fibrin in the chorionic plate. Stage 2, chorioamnionitis is characterized by neutrophilic migration into the fibrous chorion and amniotic connective tissue. Stage 3, chorioamnionitis with amnion necrosis, is characterized by presence of neutrophil karyorrhectic debris, asbcess formation, and thickened basement membranes. MIR is classically seen in the presence of ascending infection by Group B. \textit{Streptococcus}, \textit{E. coli} and other pathogenic bacteria \cite{romero2016clinical2, romero2019evidence}. However, organisms are not always identified and some authorities argue that MIR may be the result of sterile inflammation. \cite{zaidi2020clinical, romero2014prevalence, kim2015acute, lagodka2022fetal,roberts2012acute,oh2017twenty}   
\subsection*{Immediate and long-term implications of MIR}
MIR is associated with several adverse health outcomes. Bacteria causing MIR may enter the maternal or fetal circulation, causing puerperal  fever or early onset neonatal sepsis, respectively. When MIR co-occurs with FIR, there is an increased risk of neonatal death \cite{lau2005chorioamnionitis, lynch2018role}. MIR2 is a risk factor for recurrent wheeze \cite{kumar2008prematurity, mcdowell2016pulmonary}, asthma \cite{kumar2008prematurity, getahun2010effect, dessardo2019paths}, and chronic lung disease \cite{dessardo2019paths}. MIR has also been associated with lower mental development index \cite{xiao2018maternal}, and increased risk of autism spectrum disorder \cite{straughen2017association}. In a small case-control study, severe MIR has also been linked to cerebral palsy albeit indirectly \cite{redline2000relationship}.
\subsection*{Challenges in analysing MIR}
The stages of MIR are broadly defined with significant variability in how different pathologists classify these patterns and the cut-off between different stages \cite{goldstein2020maternal, redline2022interobserver}.  In a study by Redline et al. \cite{redline2022interobserver}, showed a moderate kappa agreement of $0.58$ for the diagnosis of no MIR (MIR0) vs. any MIR (MIR1-3). An older study using a previous definition of histologic chorioamnionitis showed a kappa of 0.72 \cite{simmonds2004intraobserver}. A study using secondary review of generalist pathologist diagnoses by an expert placental pathologist showed a kappa for any MIR of 0.6 \cite{ernst2023comparison}. This variability creates challenges for analysis. This may explain the relatively loose association between histologic chorioamnionitis and clinical chorioamnionitis or markers of maternal systemic inflammation \cite{rallis2022clinical, oh2017twenty, lagodka2022fetal, maki2022histological}. 

% definitions are inherently subjective
% different pathologists different threshold interobserver variablity modest kappa
% Moderately correlated and not predictive of neobatal sepsis

\subsection*{Machine Learning and Multiple Instance Learning}
Machine Learning along with digital pathology allows investigations to be conducted with larger datasets \cite{chang2019artificial, cui2021artificial, song2023artificial}. Consequently, there have been several studies conducted in recent years utilizing machine learning to build tools for pathological diagnoses, and analyse digital pathology datasets. These include both local-level tasks like nuclei segmentation, and global-level tasks prostate cancer detection \cite{srinidhi2021deep}.

Whole slide images are up to 100,000 x 60,000 pixels -- too large for efficient computation. The earliest studies in computational pathology randomly selected patches from WSI in the hope that they would carry diagnostic material from the slide. Then convolutional neural networks (ResNet, EfficientNet) or attention-based models (ViT) were finetuned on the resultant dataset \cite{laleh2022benchmarking}. Later, \textbf{multiple instance learning} (MIL) based methods were proposed, that take all the patches into account, and learn to ascribe importance to patches. The importance scores can be calculated using probability of belonging to the positive class \cite{campanella2019clinical}, as a proxy for selecting highly important patches. Another approach was proposed in \cite{lu2021data}, where attention mechanism was used to force the model to ascribe importance to patches during classification. These attention scores were calculated based on features extracted using some pretrained model e.g., ResNet50, instead of using raw patches directly. Recently, several groups have released pathology foundation models. Unlike ResNet or EfficientNet, these models use transformer architecture and have been trained on tens of thousands to millions of slides \cite{filiot2023scaling, chen2024uni,zimmermann2024virchow}. These models show superior performance on benchmarks. However, the training sets lack placental slides and benchmarks are nearly all on detection, classification, and mutation identification of neoplasia. 
% \textcolor{blue}{ Ramin's paper (in review/arxiv \cite{nateghi2024})} 

Recently, several studies have been conducted employing machine learning and MIL for predicting and analyzing placenta \cite{marletta2023application}. Chen et al. \cite{chen2020ai} used CNNs to design a system for morphological characterization, and clinically meaningful feature analysis of placentas from photos. Zhang et al. \cite{zhang2020multi} propose a Cycle-GAN along with attention module and saliency constraint to enable cross-domain image segmentation by translating target domain placenta pictures (from 1 hospital) into source domain (a different hospital), and adapting a pretrained segmentation model to segment them.
Clymer et al. \cite{clymer2020decidual} designed multiresolution CNNs to classify desidual vasculopathy in placental membranes. Mobadersany et al. \cite{mobadersany2021gestaltnet} used MIL framwork to improve prediction of gestational age from placental WSI. Andreasen et al. \cite{andreasen2023multi} proposed deep learning method for placenta segmentation from obstetric ultrasound. Goldstein et al. \cite{goldstein2023machine} used the MIL framework for classification of placental villous infarction, perivillous fibrin deposition, and intervillous thrombus. Patnaik et al. \cite{patnaik2024automated} used a pretrained ResNet-18 for feature extraction from placental histopathology images, and classify maternal vascular malperfusion.
Irmakci et al. \cite{irmakci2024tissue} investigated the challenges posed to Machine Learning models by presence of tissue contaminants e.g., floaters in WSIs. Studies have also been conducted to classify cell and regions in the placental disc \cite{ferlaino2018towards, vanea2022happy}

Recently, Chou et al. \cite{teresa2024} used machine learning and other quantitative techniques to characterize maternal inflammatory response (MIR) in placental membranes.
% \textcolor{blue}{Other MIR related papers}

In this study, we investigate MIR stage prediction from WSI. 

\section{Methods}
\subsection*{Dataset: Patients and Placentas}
The study was approved by our Institutional Review Board as STU00214052 and operated under waiver of consent. Inclusion criteria were singleton placentas examined at our institution between 2010 and 2024. We selected cases with and without MIR for retrieval and scanning.  Slides were digitized on a Leica GT450 scanner with a 40× objective magnification (0.263 $\mu m$  per pixel). Patient demographic, laboratory, and placental pathology information were extracted from our electronic data warehouse.  Slides were linked to pathology reports via optical character recognition of the labels with human review. MIR stage involving the membranes was identified from reports using regular expressions \cite{van1995python}. Unfortunately, our site does not grade MIR. Highest fever ($T_{max}$) was defined as the highest maternal temperature recorded in the 72 hours before delivery. White blood cell count (WBC) was defined as the highest maternal white blood cell count in samples collected in the 72 hours before delivery.
 
\paragraph{Data processing for MIR prediction}
We simplified the MIR stage classification problem by combining data from MIR0 and MIR1 into a one class, and MIR2 and MIR3 into another.  Patches of size $224 \times 224$ with no overlap were extracted at 20x from whole slide images of placental membranes from each patient. We used three different feature extractors. EfficientNetV2S \cite{tan2021efficientnetv2} was trained using the ImageNet dataset of everyday objects (non-medical) \cite{deng2009imagenet}. Phikon \cite{filiot2023scaling} and UNI \cite{chen2024uni} are pathology foundation models, trained on diverse datasets of histology images, though not including placenta. These features were stored in 'tfrecords' files for the aggregation and classification step in MIL pipeline.

Dataset was split into training, validation and testing splits, where $20\%$ of the data (678 samples) were held-out as the test set, and the training (2436 samples) and validation set (271 samples) were used for training and hyperparameter optimization respectively.

\paragraph{Data processing for white blood cell count and fever prediction}

For white blood cell count and $T_{max}$ prediction, we used UNI and Phikon for extracting features which were then aggregated using learned attention scores, and used by the regression network for prediction. The data split between training, validation, and test was created in a similar manner as for MIR prediction. However, there are fewer data point available with WBC (1320 training, 142 validation, and 370 test samples)  and $T_{max}$ (2256 training, 251 validation, and 632 test samples) information.   
% \begin{figure}[t]
%  % Caption and label go in the first argument and the figure contents
%  % go in the second argument
% \floatconts
%   {fig:fig1}
%   {\caption{Model predictions of segmentation results of three segmented labels produced by Re-DiffiNet on BraTS2023 dataset. First row shows the four input contrasts, the Second row shows our model predictions, and the third row shows the ground truth labels.  }}
%   {\includegraphics[width=0.5\linewidth]{Figure1.PNG}}
% \end{figure}

\subsection*{Machine Learning Models}
We used an attention-based multiple instance learning architecture similar to  \cite{lu2021data}. The architecture splits information processing into two stages- First stage involves patch-level feature extraction using a pretrained network. The pretrained network filters information from patches, that is available during stage 2 i.e., aggregation
for the final prediction. Thus, the representations learnt by feature extractor model can play a crucial role in the final predictions.
Stage 2 involves aggregation of patch-level feature to make the final prediction. The architecture uses attention to weigh different patch features. These attention weights are themselves learned during training.

% \begin{figure*}[t]
%  % Caption and label go in the first argument and the figure contents
%  % go in the second argument
% % \floatconts
%  {\includegraphics[width=\linewidth]{model_architecture_1.png}}
%   \label{FIG:fig1}
%   {\caption{Model architectures for MIR stage classification (Top), and white blood cell and Fever prediction (Bottom).}} 
% \end{figure*}

\begin{figure*}
	\centering
	\includegraphics[width=\linewidth]{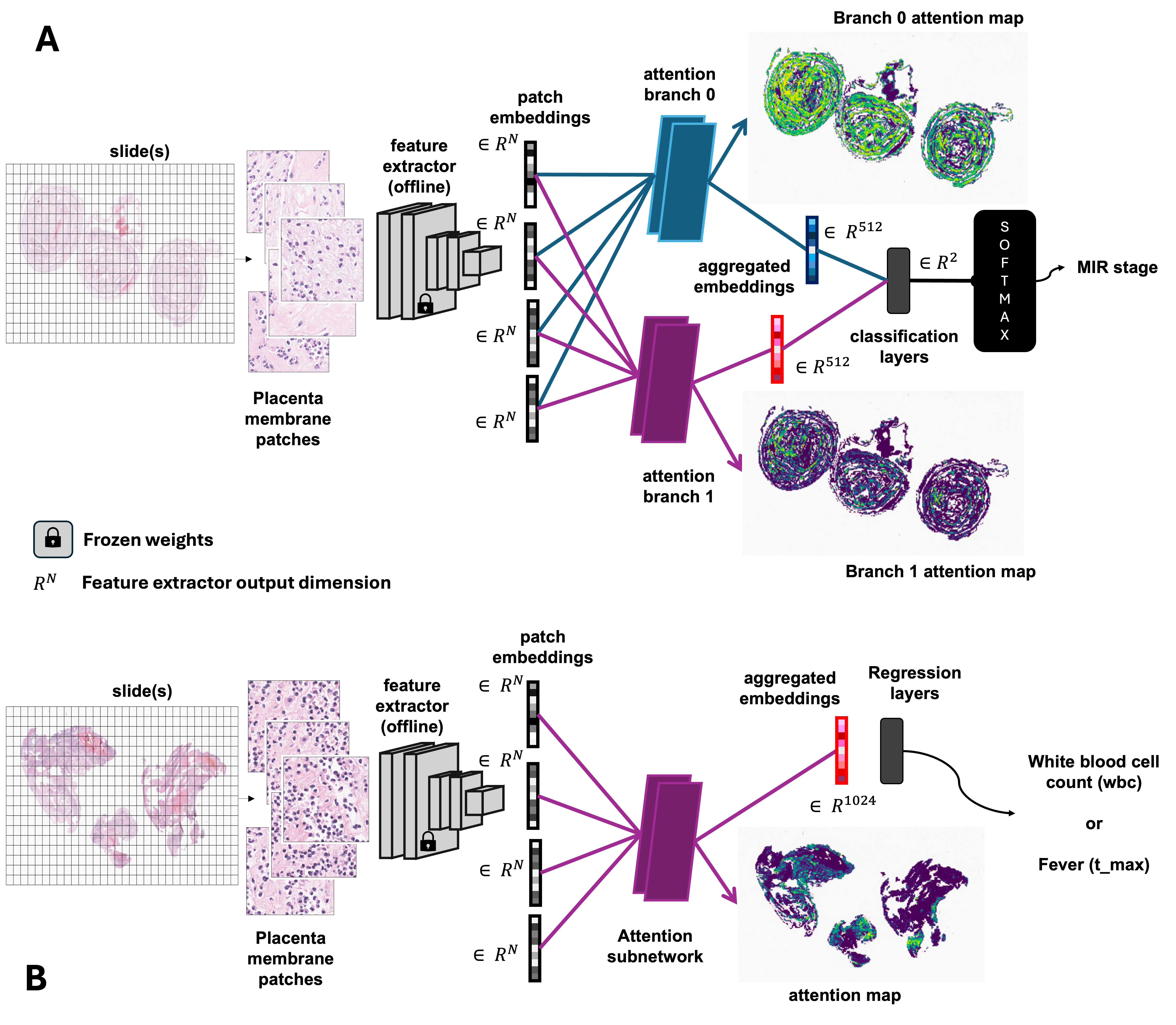}
	\caption{(A) Model architecture for MIR stage classification. Each MIR class has a corresponding set of attention weights for the patch embeddings \cite{lu2021data}. The aggregated embeddings for each class are processed by fully connected classification layers followed by a softmax layer to convert to class probabilities. The class with highest probability is the model prediction. Attention maps are visualized by coloring each patch based on corresponding attention weight.
 (B)  Model architecture for white blood cell count (WBC) and maternal highest temperature ($T_{max}$) prediction 
 }
	\label{FIG:fig1}
\end{figure*}%https://www.overleaf.com/project/66fad525f90032af61aa6478/file/66faddd6e414536130af1e15
\paragraph{Foundation models for feature extraction:} As a first step for the MIL pipeline, we extracted patch-level features from each slide using different feature extractors. We used EfficientNet-v2s \cite{tan2021efficientnetv2}, trained on ImageNet dataset (a non-medical dataset) \cite{deng2009imagenet}, which served as our model performance baseline. We investigated and compared features from two pathology foundation models- 1) Phikon \cite{filiot2023scaling}, and 2) UNI \cite{chen2024uni}, against the baseline.  
\paragraph{Loss functions:} For MIR stage prediction we used hinge loss. We also investigated effect of class weighting along with these loss functions.
For white blood cell count prediction and fever prediction, we used mean squared error (mse), and weighted mean squared error (wmse). For weighted mean squared error calculation, we first split the ground truth values into $10$ bins, then assigned weights to these bins which are inversely proportional to the number of samples in the corresponding bin. The mean squared error corresponding to each sample is then weighted according to the bin its ground truth belongs to.

\subsection*{Evaluation Metrics}
For MIR prediction, we used area under the receiver operator characteristic curve (auroc), Balanced Accuracy, Mathew's Correlation Coefficient (mcc), and Cohen's Kappa ($\kappa$) to evaluation the classification task. To evaluate white blood cell count prediction and fever prediction, we used mean squared error (MSE), mean absolute error (MAE), and $R^2$-score.

\section{Results}
Table \ref{TAB:1} summarizes the information about patients. Patients with MIR0,1 and MIR2,3 were of similar maternal age at delivery, and had a similar gestational age, although with a higher variation. Patients with MIR2,3 were more likely to be nulliparous, and less likely to have diabetes and hypertension in pregnancy, than patients with MIR0,1. \textcolor{red}{}

\begin{table*}%[width=.9\linewidth,cols=4,pos=h]
\caption{The median (IQR) values for maternal age and gestational age are shown below, along with the count (percentage) for parous, diabetes, and hypertension. p-values are reported from Mann-Whitney U rank test for Maternal and Gestational Age, and Chi-squared test for Parous, Diabetes, and Hyptension. }\label{TAB:1}

\begin{tabular*}{\tblwidth}{@{} LCCC@{} }
\hline
\textbf{value} & \textbf{MIR 0,1 (n = 2786)}    & \textbf{MIR 2, 3 (n = 599)}    & \textbf{p\_{mir01}\_{mir23}} \\
\hline
Maternal Age (yrs.) & 33 (30 - 36) & 32 (28 - 36) & 6.74E-07
\\
Gestational Age (wks.)  & 37 (34 - 39) & 37 (27 - 39) & 0.000275\\
Parous & 1352 (0.49) & 221 (0.38) & 6.15E-05\\
Diabetes in Pregnancy & 333 (0.12) & 56 (0.09) & 0.088\\
Hypertension in Pregnancy & 764 (0.27) & 80 (0.13) & 3.97E-10\\
\hline
\end{tabular*}
\end{table*}

% \begin{table*}[t]
%  % The first argument is the label.
%  % The caption goes in the second argument, and the table contents
%  % go in the third argument.
% % \floatconts
%     \label{TAB:1}%
%     \caption{\textcolor{red}{The median (IQR) values for maternal age and gestational age are shown below, along with the count (percentage) for parous, diabetes, and hypertension}}%
%     {\begin{tabular}{|l|c|c|c|}
%     \hline
%     \textbf{value} & \textbf{MIR 0,1 (n = 2786)}    & \textbf{MIR 2, 3 (n = 599)}    & \textbf{p\_{mir01}\_{mir23}} \\
%     \hline
%     Maternal Age (yrs.) & 33 (30 - 36) & 32 (28 - 36) & \\
%     \hline
%     Gestational Age (wks.)  & 37 (34 - 39) & 37 (27 - 39) & \\
%     \hline
%     Parous & 1352 (0.49) & 221 (0.38) & \\
%     \hline
%     Diabetes in Pregnancy & 333 (0.12) & 56 (0.09) & \\
%     \hline
%     Hypertension in Pregnancy & 764 (0.27) & 80 (0.13) & \\
%     \hline
%     \end{tabular}}
% \end{table*}

\subsection*{MIR stage prediction: Balanced accuracy, MCC, Cohen's kappa}
Table \ref{tab:results2} shows auroc, balanced accuracy, Mathew's correlation coefficient (mcc), and Cohen's kappa ($\kappa$) values for EfficientNet, Phikon, and UNI features. These models achieved a balanced accuracy and Cohen's Kappa ($\kappa$) of $>83.7\%$, and $>0.724$ respectively.

\begin{table*}%[width=.9\linewidth,cols=4,pos=h]
\caption{Comparison of performance between different feature extractors in terms of auroc, balanced accuracy, mcc, and Cohen's Kappa, specificity, and sensitivity. Values reported are for the test set (N=676).}\label{tab:results2}
\begin{tabular*}{\tblwidth}{@{} LCCCCCC@{} }
\hline
& \textbf{AUROC}    & \textbf{Balanced Acc.}    & \textbf{MCC}    & \textbf{$\kappa$} & \textbf{Specificity} & \textbf{Sensitivity} \\
\hline
EfficientNet & 0.837 & 83.7 \% & 0.729 & 0.724 & \textbf{0.97} & 0.70\\
Phikon  & \textbf{0.885} & \textbf{88.5 \%} & \textbf{0.772} & \textbf{0.772} & 0.96 & \textbf{0.81}\\
UNI & 0.872 & 87.2 \% & 0.751 & 0.751 & 0.96 & 0.79\\
\hline
\end{tabular*}
\end{table*}

% \begin{table*}[t]
%  % The first argument is the label.
%  % The caption goes in the second argument, and the table contents
%  % go in the third argument.
% \floatconts
% \label{tab:results2}%
% {\caption{Comparison of performance between different feature extractors in terms of auroc, balanced accuracy, mcc, and Cohen's Kappa, specificity, and sensitivity. Values reported are for the test set N=676).}}%
% {\begin{tabular}{|l|c|c|c|c|c|c|}
% \hline
%  & \textbf{AUROC}    & \textbf{Balanced Acc.}    & \textbf{MCC}    & \textbf{$\kappa$} & \textbf{Specificity} & \textbf{Sensitivity} \\
% \hline
% EfficientNet & 0.837 & 83.7 \% & 0.729 & 0.724 & \textbf{0.97} & 0.70\\
% \hline
% Phikon  & \textbf{0.885} & \textbf{88.5 \%} & \textbf{0.772} & \textbf{0.772} & 0.96 & \textbf{0.81}\\
% \hline
% UNI & 0.872 & 87.2 \% & 0.751 & 0.751 & 0.96 & 0.79\\
% \hline
% \end{tabular}%
% }
% \end{table*}

\begin{figure}
	\centering
	\includegraphics[width=1\linewidth]{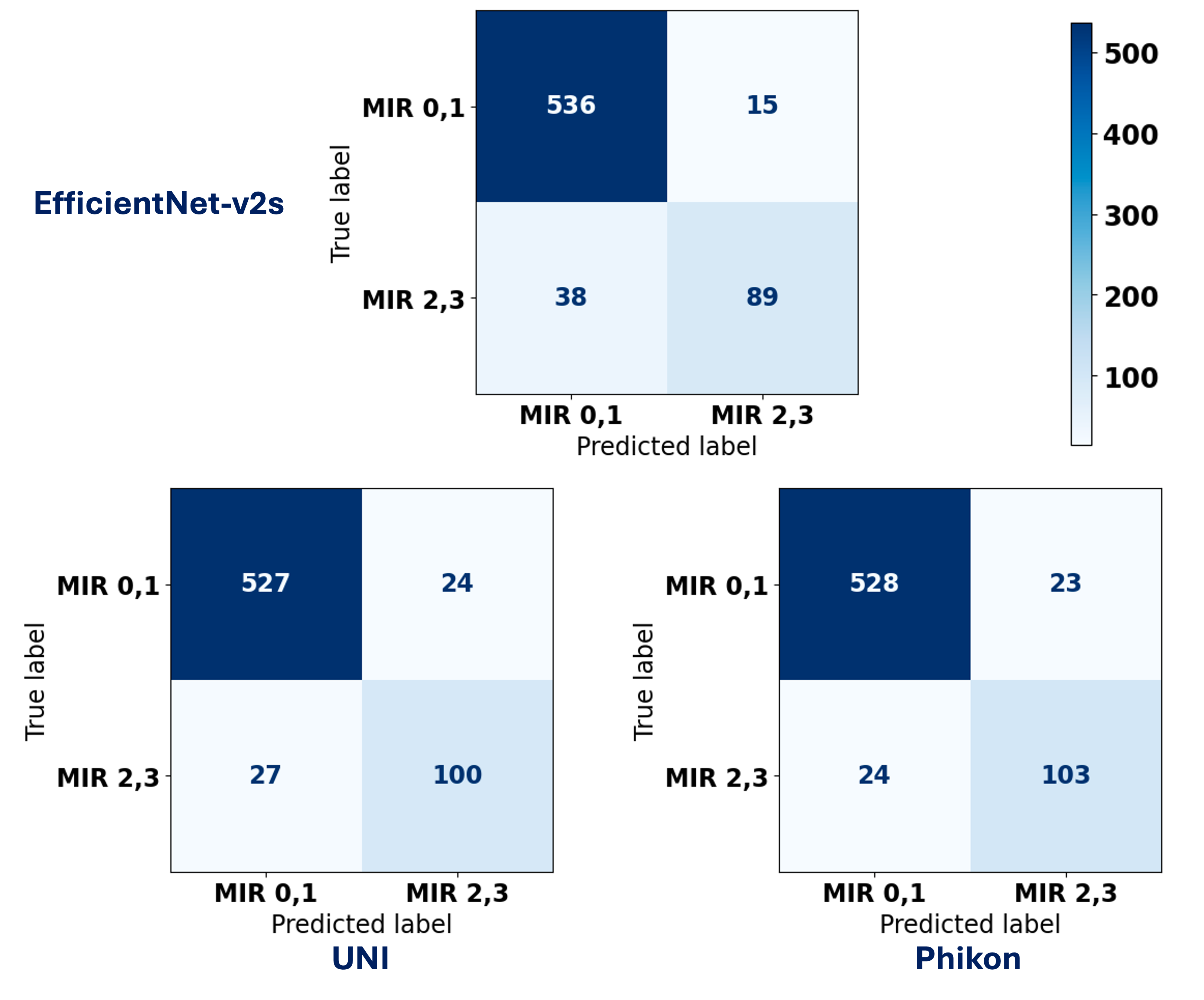}
	\caption{Test set confusion matrices for EfficientNet (top), UNI (bottom-left), Phikon (bottom-right).}
	\label{FIG:fig2}
\end{figure}

% \begin{figure}[t]
%  % Caption and label go in the first argument and the figure contents
%  % go in the second argument
% % \floatconts
%   {\includegraphics[width=1\linewidth]{corrected_confusionmatrix_comparison.png}}
%   \label{FIG:fig2}
%   {\caption{Test set confusion matrices for EfficientNet (left), UNI (center), Phikon (right).}}
% \end{figure}

\subsection*{Comparison between feature extractors}
Table \ref{tab:results2} shows a comparison between a feature extractor trained on ImageNet (EfficientNet-v2s), and 2 pathology foundation models (UNI and Phikon). EfficientNet is prone to making more false negative predictions compared to UNI and Phikon, and is worse at identifying MIR2,3. We found that both the foundation models perform significantly better than EfficientNet in terms of balanced accuracy. In addition, we found that Phikon is marginally better than UNI.

\subsection*{Visualizing slide-level and patch-level features} 
\begin{figure}
	\centering
	\includegraphics[width=\linewidth]{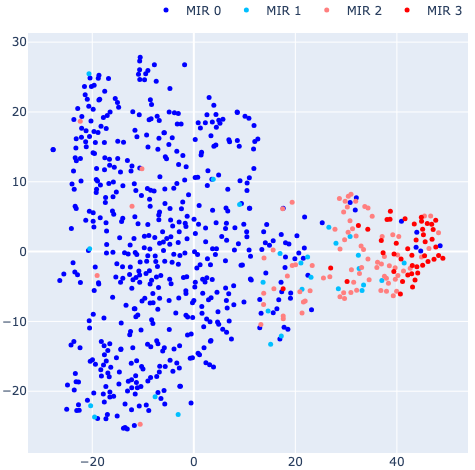}
	\caption{t-SNE space of aggregated features from Phikon in the test set. Colors represent the true class. MIR0 is shown in dark blue, MIR1 in light blue, MIR2 in light red, and MIR3 in dark red.}
	\label{FIG:fig3}
\end{figure}

Figure \ref{FIG:fig3} a the t-SNE visualization of slide-level features from the test set, aggregated by the pooling layer of the Phikon MIL model. We see slides being separated into two major clusters. Data points with MIR0 and MIR1 are shown in dark blue and light blue respectively. MIR2 and MIR3 shown in red, and light red respectively.  The aggregated embeddings show definite, but imperfect separation between the MIR0,1 and MIR2,3 slides. The embeddings also show a gradient with MIR0 slides (dark blue) farthest to the left and MIR3 slides (bright red) mostly to the right. MIR1 and MIR2 are placed in between, with MIR1 being closer to MIR0 than MIR3, while MIR2 is closer to MIR3. 

\begin{figure*}
	\centering	\includegraphics[width=1\linewidth]{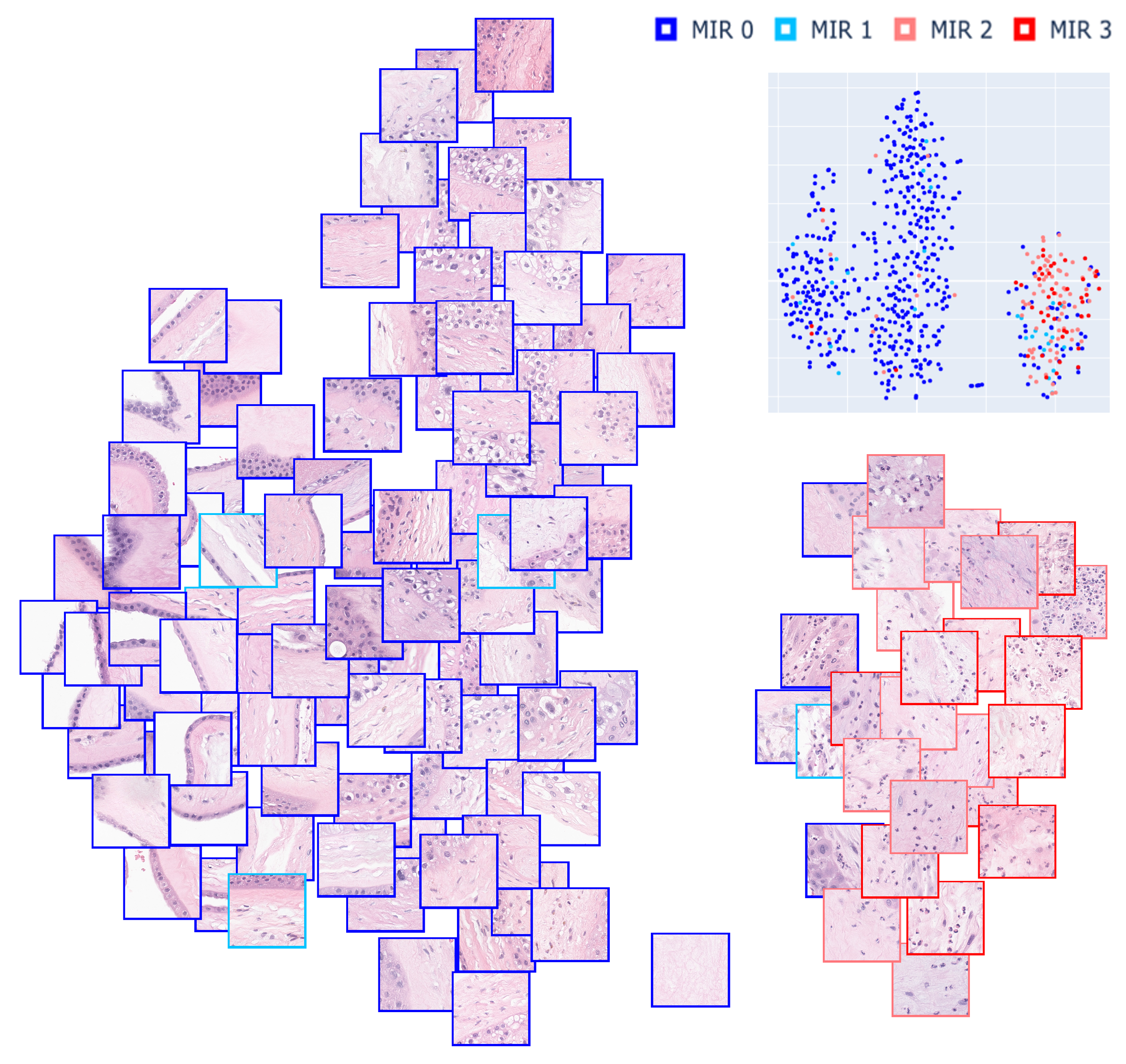}
	\caption{t-SNE embeddings of top-1 attended patches for attention branch 1 (See model architecture in Fig. \ref{FIG:fig1}), from each slide in the test set. Image patches corresponding to randomly sampled patches are shown with patching outlines highlighting the original diagnosis (MIR Stage) of corresponding WSIs. Blue patches are largely normal placenta or some neutrophil infiltration. The red patches show stroma with greater neutrophil infiltration. All the embeddings are shown in the inset.}
	\label{FIG:fig4}
\end{figure*}

Figure \ref{FIG:fig4} shows t-SNE visualization for top-1 attention patches from all the wsi. The weighted average featuer vectors encode information relevant to the MIR stage diagnosis, as reflected by separate blue and red clusters for MIR0,1 and MIR2,3 respectively. 

% \begin{figure}[t]
%   \centering  {\includegraphics[width=0.8\linewidth]{phikon_corrected_testsetallclasses_agg_emb.png}}
%   \label{FIG:fig3}
%   {\caption{TSNE space of aggregated features from Phikon in the test set. Colors represent the true class. MIR0 is shown in dark blue, MIR1 in light blue, MIR2 in light red, and MIR3 in dark red.}}
% \end{figure}

% \begin{figure*}[t]
%  % Caption and label go in the first argument and the figure contents
%  % go in the second argument
% % \floatconts
%  {\includegraphics[width=0.87\linewidth]{corrected_phikontop1_allbranches.png}} 
%   \label{FIG:fig4}
%   {\caption{TSNE space of top-1 attended patches for (a) attention branch 0, and (b) attention branch 1, from each slide in the test set. The corresponding tsne embeddings are shown in the inset. Inset outlines show the true class of corresponding WSIs.}}
%     % \begin{subfigure}
%     % \centering        {\includegraphics[width=\linewidth]{phikontop1_branch0.png}}      \end{subfigure}%
    
%     % \begin{subfigure}
%     % \centering        {\includegraphics[width=\linewidth]{phikontop1_branch1.png}}      \end{subfigure}%
      
% \end{figure*}

\subsection*{Investigating model failure cases: False positives and negatives.}

\begin{figure}
    \centering
    \includegraphics[width=\linewidth]{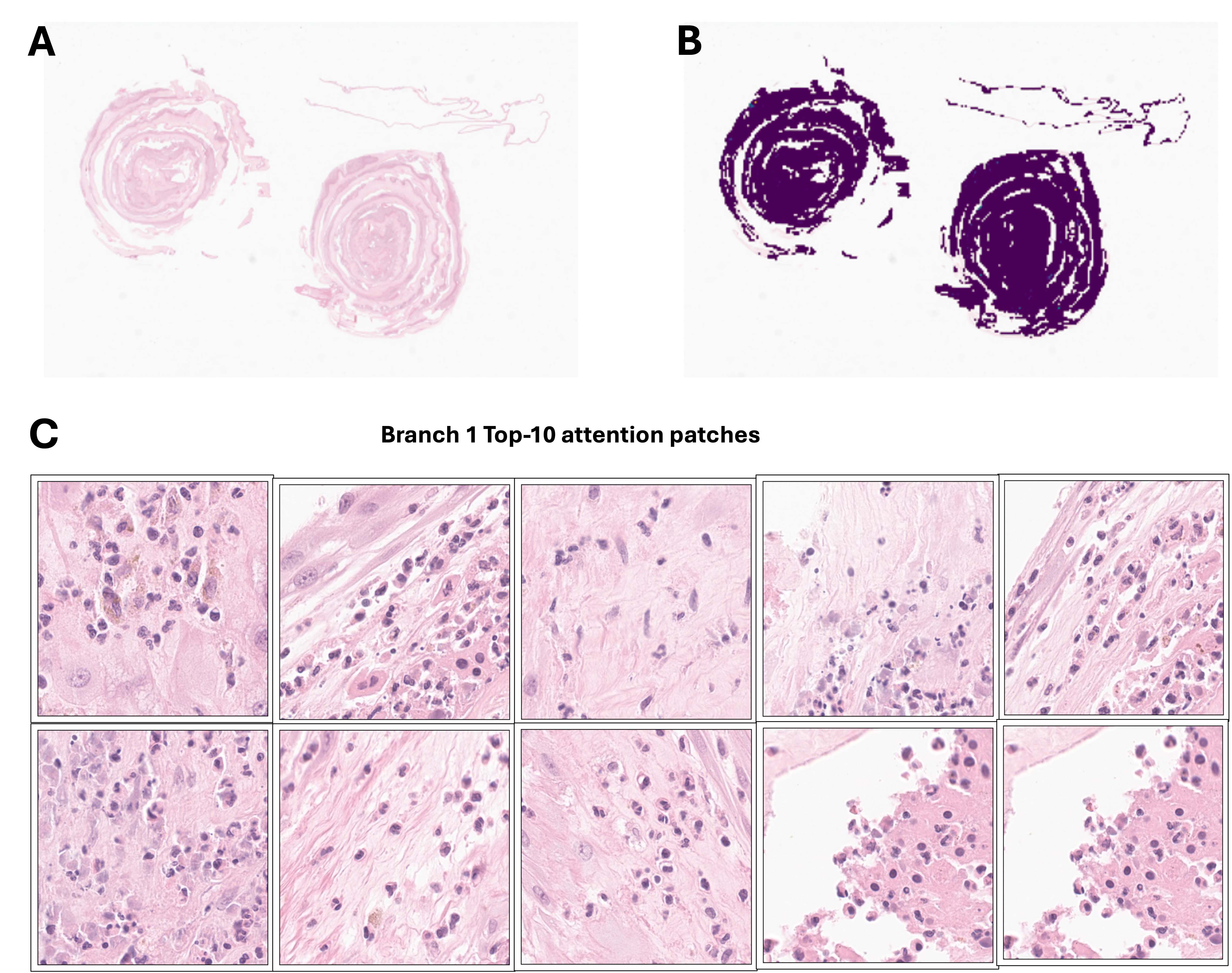}
    \caption{An example of a false positive. (A) WSI, (B) Attention Heatmap, (C) Top-10 attention patches from attention branch 1 (See model architecture in Fig. \ref{FIG:fig1}).
    This is a case with severe MIR1 with decidual necrosis, but does not meet criteria for MIR2.}
    \label{FIG:fig5}
\end{figure}

\begin{figure}
	\centering
	\includegraphics[width=\linewidth]{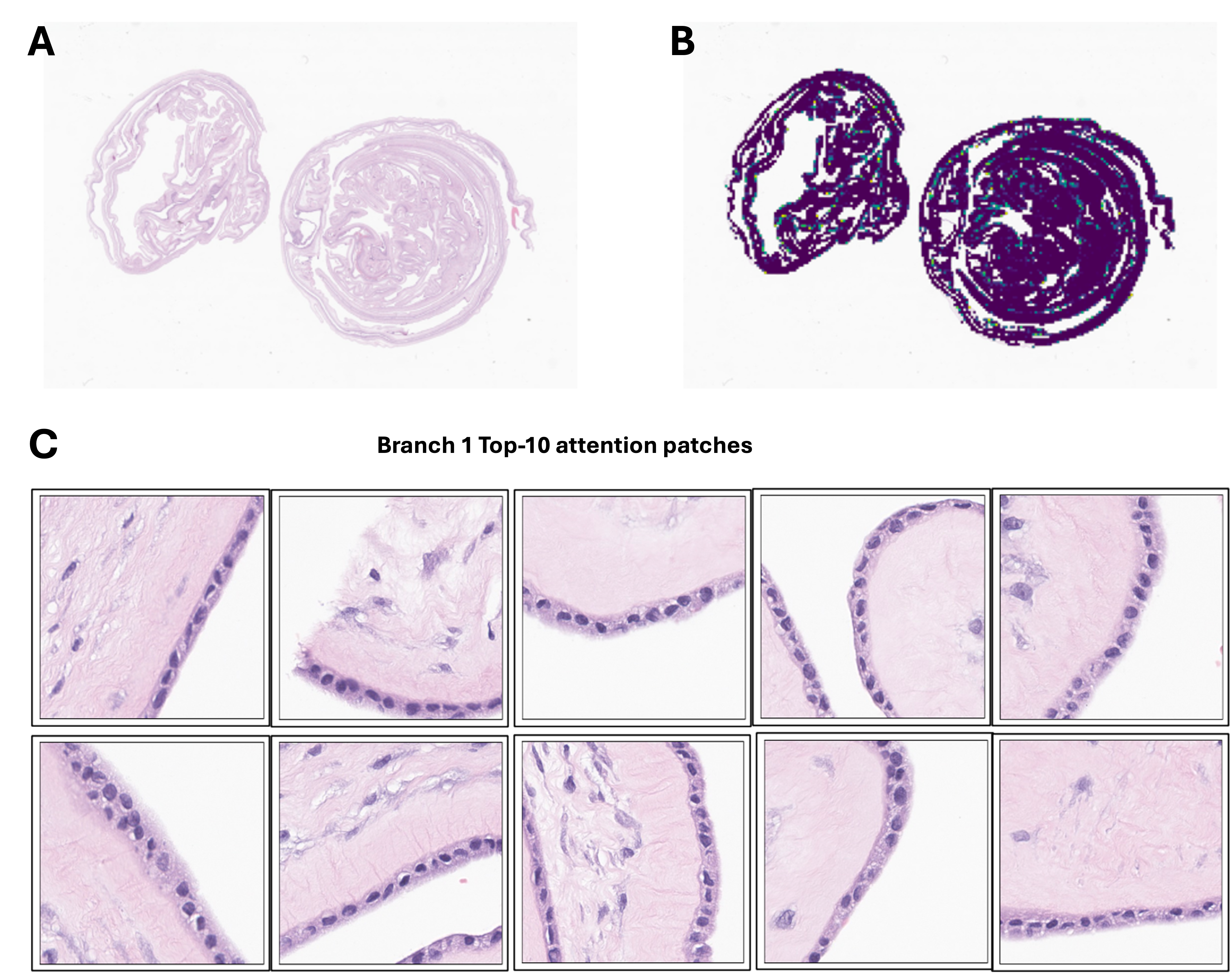}
	\caption{An example of false negative. (A) WSI, (B) Attention Heatmap, (C) Top-10 attention patches from attention branch 1 (See model architecture in Fig. \ref{FIG:fig1}). Review of high attention patches showed amnion with activated-appearing macrophages. Manual review of the whole slide showed scant inflammation, which may not be classified as MIR2 by all observers.}
	\label{FIG:fig6}
\end{figure}

To understand how the model functions or malfunctions, we  investigated a few cases where the model gives false-positive or false-negative results. Specifically, we looked at the extreme false positive (rightmost blue points in Figure \ref{FIG:fig3}), and extreme false negative (leftmost red points in Figure \ref{FIG:fig3}). The attention heatmap, and top-10 attention patches for a false positive and false negative are shown in Fig. \ref{FIG:fig5} and \ref{FIG:fig6} respectively.

% \begin{figure}[t]
%  % Caption and label go in the first argument and the figure contents
%  % go in the second argument
% % \floatconts

% {\includegraphics[width=\linewidth]{oixcggwn_7_FP_.png}
%   \label{FIG:fig5}
%   {\caption{An example of false positive. This is a case with high-grade MIR1, but does not meet criteria for MIR2. }}
% }
% \end{figure}

% \begin{figure}[t]
%  % Caption and label go in the first argument and the figure contents
%  % go in the second argument
% % \floatconts

%   {\includegraphics[width=\linewidth]{kykujfja_1_FN_.png}
%   }
%   \label{FIG:fig6}
%   {\caption{An example of false negative. This is a case with some decidual inflammation, but is an edge case and not a strong example of MIR2. }}
% \end{figure}

\subsection*{White blood cell count prediction}
Table \ref{tab:resultswbc} shows performance of UNI and Phikon-based regression networks for white blood cell prediction. UNI performs slightly better than Phikon in terms of RMSE, MAE, and $R^2$-score values. However, both the models show weak correlation between predicted and actual values of WBC.

% \begin{figure}[t]
%  % Caption and label go in the first argument and the figure contents
%  % go in the second argument
% % \floatconts

%   {\includegraphics[width=\linewidth]{wbc_combined.png}}
%   \label{fig:wbc}
%   {\caption{UNI predicts white blood cell count (wbc) with a R2 of 0.071 and rmse of 5.370. Phikon predicts wbc with a R2 of 0.043 and rmse of 7.744. (See Table \ref{tab:resultswbc})}}
% \end{figure}

\begin{figure}
	\centering
	\includegraphics[width=\linewidth]{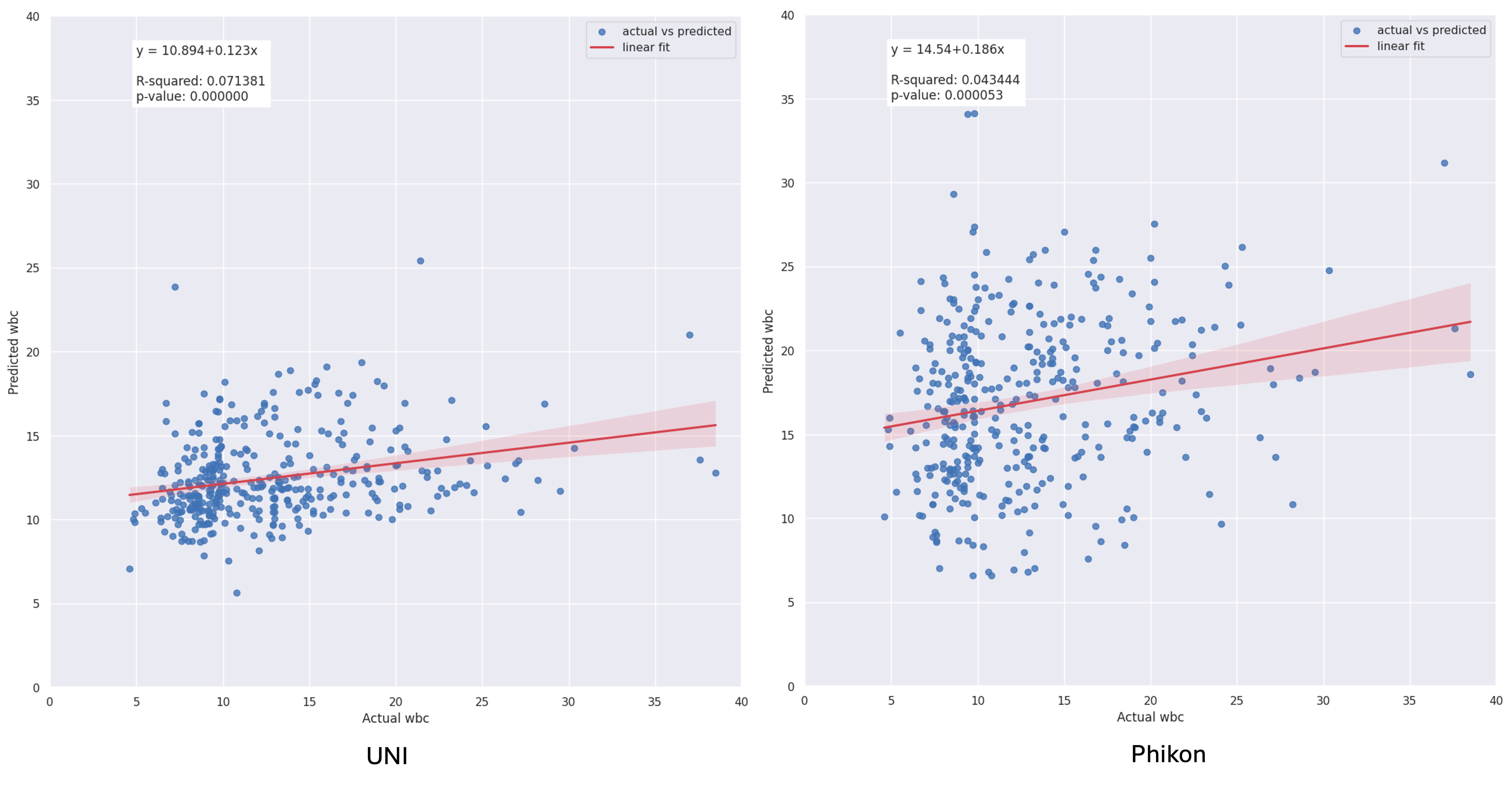}
	\caption{UNI predicts white blood cell count (WBC) with an $R\sup{2}$ of 0.071. Phikon predicts WBC with a R2 of 0.043 (See Table \ref{tab:resultswbc})}
	\label{fig:wbc}
\end{figure}

\subsection*{Maternal highest temperature prediction}
Table \ref{tab:resultswbc} shows performance of UNI and Phikon-based regression networks for fever prediction ($T_{max}$). UNI performs slightly better than Phikon in terms of RMSE, MAE, and $R^2$-score. Both the models show mild correlation between predicted and actual values of $T_{max}$.

% \begin{figure}[t]
%  % Caption and label go in the first argument and the figure contents
%  % go in the second argument
% % \floatconts
%   {\includegraphics[width=\linewidth]{tmc_combined.png}}
%   \label{fig:tmc}
%   {\caption{UNI predicts maximum fever temperature ($T_{max}$) with a R2 of 0.396 and rmse of 1.021. Phikon predicts $T_{max}$  with a R2 of 0.291 and rmse of 1.125. (See Table \ref{tab:resultswbc}) }}
% \end{figure}

\begin{figure}
	\centering
	\includegraphics[width=\linewidth]{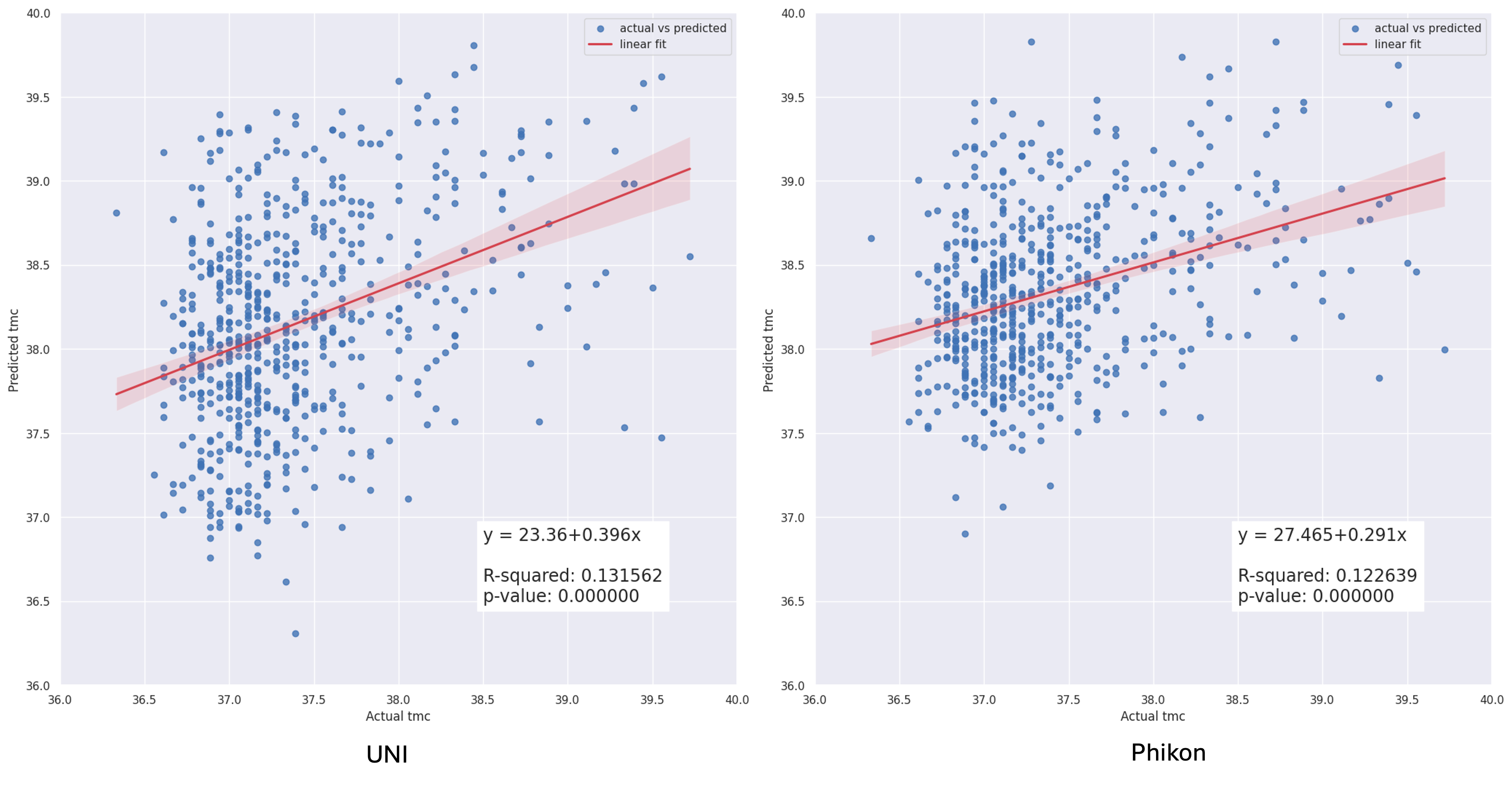}
	\caption{UNI predicts maximum fever temperature ($T_{max}$) with a R2 of 0.396 and rmse of 1.021. Phikon predicts $T_{max}$  with a R2 of 0.291 and rmse of 1.125. (See Table \ref{tab:resultswbc}) }
	\label{fig:tmc}
\end{figure}

\begin{table}%[width=.9\linewidth,cols=4,pos=h]
\caption{Performance for white blood cell count (WBC) prediction and fever ($T_{max}$) prediction using UNI features}\label{tab:resultswbc}
\begin{tabular*}{\tblwidth}{@{} CCCCCC@{} }
\hline
value & Model & \textbf{RMSE}      &  \textbf{MAE}      &  \textbf{$R^2$}      & \textbf{slope} \\
\hline
\multirow{2}{*}{WBC}         & \multicolumn{1}{l}{UNI}                   & \textbf{5.370}  & \textbf{4.045}  &   \textbf{0.071}  &  0.123   \\   
                                   & \multicolumn{1}{l}{Phikon}   & 7.744  & 6.392  & 0.043  &  \textbf{0.186}  \\ \hline
\multirow{2}{*}{$T_{max}$}         & \multicolumn{1}{l}{UNI}                   & \textbf{1.021}  & \textbf{0.856}  &  \textbf{0.132} & \textbf{0.396}  \\   
                                   & \multicolumn{1}{l}{Phikon}   & 1.125  & 0.999 & 0.123 & 0.291   \\ \hline
\end{tabular*}
\end{table}

% \begin{table}[t]
%  % The first argument is the label.
%  % The caption goes in the second argument, and the table contents
%  % go in the third argument.
% % \floatconts
% \label{tab:resultswbc}%
% {\caption{Performance for white blood cell count (wbc) prediction and fever ($T_{max}$) prediction using UNI features}}%
% {\begin{tabular}{c|c|cccc}
%     \hline
%     value & Model & \textbf{RMSE}      &  \textbf{MAE}      &  \textbf{$R^2$}      & \textbf{slope} \\ \hline
%     \multirow{2}{*}{wbc}         & \multicolumn{1}{l|}{UNI}                   & \textbf{5.370}  & \textbf{4.045}  &   \textbf{0.071}  &  0.123   \\   
%                                    & \multicolumn{1}{l|}{Phikon}   & 7.744  & 6.392  & 0.043  &  \textbf{0.186}  \\ \hline
%     \multirow{2}{*}{$T_{max}$}         & \multicolumn{1}{l|}{UNI}                   & \textbf{1.021}  & \textbf{0.856}  &  \textbf{0.132} & \textbf{0.396}  \\   
%                                    & \multicolumn{1}{l|}{Phikon}   & 1.125  & 0.999 & 0.123 & 0.291   \\ \hline

% \end{tabular}}
% \end{table}

\section{Discussion}
\textbf{Automated analysis of MIR}

We investigated the feasibility of predicting MIR stages (MIR0,1 and MIR2,3) from whole slide images. We found that attention-based MIL models are able to classify MIR with a balanced accuracy of up to 88.5\% with a Cohen's $\kappa$ of up to 0.772. 
Furthermore, we found that the pathology foundation models (UNI and Phikon) are both able to achieve higher performance with balanced accuracy of 87.2 \% and 88.5 \% respectively, and Cohen's Kappa $(\kappa)$ of 0.751 and 0.772 respectively, compared to ImageNet-based feature extractor (EfficientNet-v2s). This is despite the fact that both the pathology foundation models have not been pretrained on any placental data.

We also investigated prediction of white blood cell count (WBC), and fever ($T_{max}$) from WSI. We found that the regression model with the UNI feature extractor, shows a moderate correlation between the predicted and actual $T_{max}$, indicating a moderate ability to predict fever from histopathological images. Prediction of white blood cell count shows a much weaker correlation.

\textbf{Comparison between feature extractors} In this study, we found that feature from histology foundation models- UNI, and Phikon, outperform ImageNet features for MIR stage prediction, despite the foundation models not being pretrained on any placenta data. In addition, we found Phikon to be slightly better than UNI. 

\textbf{Interobserver variability}
Of the cases misclassified by our model, many showed borderline findings between MIR1 and MIR2.  This highlights the challenges of interobserver variability in building and assessing models. In a different context, we found that training on noisy labels from multiple observers  can yield expert level performance \cite{amgad2022nucls}. Future studies could show similar improvement.

\textbf{Clinical chorioamnionitis} is defined by maternal fever, fetal or maternal tachycardia, uterine tenderness, and foul discharge, and is often accompanied by elevated WBC count \cite{ajayi2022association, romero2016clinical, jung2023clinical}. As an exploratory study, we tested whether we could estimate $T_{max}$ or WBC using placental histology. We found weak, but statistically significant correlations.

\paragraph{Strengths and limitations:}
One of the major strengths of this study is that the dataset the large (N = 3,385) real-world dataset.  Further, we have investigated the use of medical imaging foundation models (Phikon and UNI)
as feature extractors in the MIL pipeline, and demonstrated that they perform better than using features extracted from model pretrained on images from everyday objects i.e., ImageNet (EfficientNet-v2s), for MIR classification, even though no placental images were used to train Phikon and UNI. We also showed that Phikon and UNI both achieve similar performance for the MIR prediction task. Thus highlighting the need for finetuning on placenta data for these models to extract finer details from placenta images, and achieve further performance gains. We investigated model's ability to predict fever from histopathological images, and found moderate correlation between predicted and actual $T_{max}$. 

One of the weaknesses of this study is that the dataset was sourced from a single site. The models might not generalize to datasets sources from other sites. Further limitations include that other maternal or fetal signs of inflammation were not considered. The rarity of neonatal sepsis poses challenges to predict it, using our data. Also, due to the nature of our dataset, long-term neonatal outcomes could not be investigated.

\section*{Contributions} Conception of the work: J.A.G., L.A.D.C., A.S.; Patient selection/scanning: J.A.G.; Technique and tool development: A.S., R.N., L.A.D.C., M.A.; Data preparation: A.S.; Experiments: A.S.; Analysis: A.S., J.A.G.; Manuscript: A.S.; Editing: A.S., J.A.G., L.A.D.C.

\section*{Declaration of competing interest}
The authors have no relevant disclosures


\bibliographystyle{cas-model2-names}

\bibliography{cas-refs}

\begin{thebibliography}{53}
\expandafter\ifx\csname natexlab\endcsname\relax\def\natexlab#1{#1}\fi
\providecommand{\url}[1]{\texttt{#1}}
\providecommand{\href}[2]{#2}
\providecommand{\path}[1]{#1}
\providecommand{\DOIprefix}{doi:}
\providecommand{\ArXivprefix}{arXiv:}
\providecommand{\URLprefix}{URL: }
\providecommand{\Pubmedprefix}{pmid:}
\providecommand{\doi}[1]{\href{http://dx.doi.org/#1}{\path{#1}}}
\providecommand{\Pubmed}[1]{\href{pmid:#1}{\path{#1}}}
\providecommand{\bibinfo}[2]{#2}
\ifx\xfnm\relax \def\xfnm[#1]{\unskip,\space#1}\fi
%Type = Article
\bibitem[{Ajayi et~al.(2022)Ajayi, Morris, Aleem, Pease, Wang, Mowes, Welles, Anday and Bhandari}]{ajayi2022association}
\bibinfo{author}{Ajayi, S.O.}, \bibinfo{author}{Morris, J.}, \bibinfo{author}{Aleem, S.}, \bibinfo{author}{Pease, M.E.}, \bibinfo{author}{Wang, A.}, \bibinfo{author}{Mowes, A.}, \bibinfo{author}{Welles, S.L.}, \bibinfo{author}{Anday, E.K.}, \bibinfo{author}{Bhandari, V.}, \bibinfo{year}{2022}.
\newblock \bibinfo{title}{Association of clinical signs of chorioamnionitis with histological chorioamnionitis and neonatal outcomes}.
\newblock \bibinfo{journal}{The Journal of Maternal-Fetal \& Neonatal Medicine} \bibinfo{volume}{35}, \bibinfo{pages}{10337--10347}.
%Type = Article
\bibitem[{Amgad et~al.(2022)Amgad, Atteya, Hussein, Mohammed, Hafiz, Elsebaie, Alhusseiny, AlMoslemany, Elmatboly, Pappalardo et~al.}]{amgad2022nucls}
\bibinfo{author}{Amgad, M.}, \bibinfo{author}{Atteya, L.A.}, \bibinfo{author}{Hussein, H.}, \bibinfo{author}{Mohammed, K.H.}, \bibinfo{author}{Hafiz, E.}, \bibinfo{author}{Elsebaie, M.A.}, \bibinfo{author}{Alhusseiny, A.M.}, \bibinfo{author}{AlMoslemany, M.A.}, \bibinfo{author}{Elmatboly, A.M.}, \bibinfo{author}{Pappalardo, P.A.}, et~al., \bibinfo{year}{2022}.
\newblock \bibinfo{title}{Nucls: A scalable crowdsourcing approach and dataset for nucleus classification and segmentation in breast cancer}.
\newblock \bibinfo{journal}{GigaScience} \bibinfo{volume}{11}, \bibinfo{pages}{giac037}.
%Type = Article
\bibitem[{Andreasen et~al.(2023)Andreasen, Feragen, Christensen, Thybo, Svendsen, Zepf, Lekadir and Tolsgaard}]{andreasen2023multi}
\bibinfo{author}{Andreasen, L.A.}, \bibinfo{author}{Feragen, A.}, \bibinfo{author}{Christensen, A.N.}, \bibinfo{author}{Thybo, J.K.}, \bibinfo{author}{Svendsen, M.B.S.}, \bibinfo{author}{Zepf, K.}, \bibinfo{author}{Lekadir, K.}, \bibinfo{author}{Tolsgaard, M.G.}, \bibinfo{year}{2023}.
\newblock \bibinfo{title}{Multi-centre deep learning for placenta segmentation in obstetric ultrasound with multi-observer and cross-country generalization}.
\newblock \bibinfo{journal}{Scientific Reports} \bibinfo{volume}{13}, \bibinfo{pages}{2221}.
%Type = Article
\bibitem[{Campanella et~al.(2019)Campanella, Hanna, Geneslaw, Miraflor, Werneck Krauss~Silva, Busam, Brogi, Reuter, Klimstra and Fuchs}]{campanella2019clinical}
\bibinfo{author}{Campanella, G.}, \bibinfo{author}{Hanna, M.G.}, \bibinfo{author}{Geneslaw, L.}, \bibinfo{author}{Miraflor, A.}, \bibinfo{author}{Werneck Krauss~Silva, V.}, \bibinfo{author}{Busam, K.J.}, \bibinfo{author}{Brogi, E.}, \bibinfo{author}{Reuter, V.E.}, \bibinfo{author}{Klimstra, D.S.}, \bibinfo{author}{Fuchs, T.J.}, \bibinfo{year}{2019}.
\newblock \bibinfo{title}{Clinical-grade computational pathology using weakly supervised deep learning on whole slide images}.
\newblock \bibinfo{journal}{Nature medicine} \bibinfo{volume}{25}, \bibinfo{pages}{1301--1309}.
%Type = Article
\bibitem[{Chang et~al.(2019)Chang, Jung, Woo, Lee, Cho, Kim and Kwak}]{chang2019artificial}
\bibinfo{author}{Chang, H.Y.}, \bibinfo{author}{Jung, C.K.}, \bibinfo{author}{Woo, J.I.}, \bibinfo{author}{Lee, S.}, \bibinfo{author}{Cho, J.}, \bibinfo{author}{Kim, S.W.}, \bibinfo{author}{Kwak, T.Y.}, \bibinfo{year}{2019}.
\newblock \bibinfo{title}{Artificial intelligence in pathology}.
\newblock \bibinfo{journal}{Journal of pathology and translational medicine} \bibinfo{volume}{53}, \bibinfo{pages}{1--12}.
%Type = Article
\bibitem[{Chen et~al.(2024)Chen, Ding, Lu, Williamson, Jaume, Chen, Zhang, Shao, Song, Shaban et~al.}]{chen2024uni}
\bibinfo{author}{Chen, R.J.}, \bibinfo{author}{Ding, T.}, \bibinfo{author}{Lu, M.Y.}, \bibinfo{author}{Williamson, D.F.}, \bibinfo{author}{Jaume, G.}, \bibinfo{author}{Chen, B.}, \bibinfo{author}{Zhang, A.}, \bibinfo{author}{Shao, D.}, \bibinfo{author}{Song, A.H.}, \bibinfo{author}{Shaban, M.}, et~al., \bibinfo{year}{2024}.
\newblock \bibinfo{title}{Towards a general-purpose foundation model for computational pathology}.
\newblock \bibinfo{journal}{Nature Medicine} .
%Type = Article
\bibitem[{Chen et~al.(2020)Chen, Zhang, Wu, Davaasuren, Goldstein, Gernand and Wang}]{chen2020ai}
\bibinfo{author}{Chen, Y.}, \bibinfo{author}{Zhang, Z.}, \bibinfo{author}{Wu, C.}, \bibinfo{author}{Davaasuren, D.}, \bibinfo{author}{Goldstein, J.A.}, \bibinfo{author}{Gernand, A.D.}, \bibinfo{author}{Wang, J.Z.}, \bibinfo{year}{2020}.
\newblock \bibinfo{title}{Ai-plax: Ai-based placental assessment and examination using photos}.
\newblock \bibinfo{journal}{Computerized Medical Imaging and Graphics} \bibinfo{volume}{84}, \bibinfo{pages}{101744}.
%Type = Article
\bibitem[{Chou et~al.(2024)Chou, Senkow, Nguyen, Patel, Sandepudi, Cooper and Goldstein}]{teresa2024}
\bibinfo{author}{Chou, T.}, \bibinfo{author}{Senkow, K.J.}, \bibinfo{author}{Nguyen, M.B.}, \bibinfo{author}{Patel, P.V.}, \bibinfo{author}{Sandepudi, K.}, \bibinfo{author}{Cooper, L.A.}, \bibinfo{author}{Goldstein, J.A.}, \bibinfo{year}{2024}.
\newblock \bibinfo{title}{Quantitative modeling to characterize maternal inflammatory response of histologic chorioamnionitis in placental membranes}.
\newblock \bibinfo{journal}{American Journal of Reproductive Immunology} \bibinfo{volume}{92}, \bibinfo{pages}{e13944}.
\newblock \URLprefix \url{https://onlinelibrary.wiley.com/doi/abs/10.1111/aji.13944}, \DOIprefix\doi{https://doi.org/10.1111/aji.13944}, \href{http://arxiv.org/abs/https://onlinelibrary.wiley.com/doi/pdf/10.1111/aji.13944}{\tt arXiv:https://onlinelibrary.wiley.com/doi/pdf/10.1111/aji.13944}.
%Type = Article
\bibitem[{Clymer et~al.(2020)Clymer, Kostadinov, Catov, Skvarca, Pantanowitz, Cagan and LeDuc}]{clymer2020decidual}
\bibinfo{author}{Clymer, D.}, \bibinfo{author}{Kostadinov, S.}, \bibinfo{author}{Catov, J.}, \bibinfo{author}{Skvarca, L.}, \bibinfo{author}{Pantanowitz, L.}, \bibinfo{author}{Cagan, J.}, \bibinfo{author}{LeDuc, P.}, \bibinfo{year}{2020}.
\newblock \bibinfo{title}{Decidual vasculopathy identification in whole slide images using multiresolution hierarchical convolutional neural networks}.
\newblock \bibinfo{journal}{The American Journal of Pathology} \bibinfo{volume}{190}, \bibinfo{pages}{2111--2122}.
%Type = Article
\bibitem[{Cui and Zhang(2021)}]{cui2021artificial}
\bibinfo{author}{Cui, M.}, \bibinfo{author}{Zhang, D.Y.}, \bibinfo{year}{2021}.
\newblock \bibinfo{title}{Artificial intelligence and computational pathology}.
\newblock \bibinfo{journal}{Laboratory Investigation} \bibinfo{volume}{101}, \bibinfo{pages}{412--422}.
%Type = Inproceedings
\bibitem[{Deng et~al.(2009)Deng, Dong, Socher, Li, Li and Fei-Fei}]{deng2009imagenet}
\bibinfo{author}{Deng, J.}, \bibinfo{author}{Dong, W.}, \bibinfo{author}{Socher, R.}, \bibinfo{author}{Li, L.J.}, \bibinfo{author}{Li, K.}, \bibinfo{author}{Fei-Fei, L.}, \bibinfo{year}{2009}.
\newblock \bibinfo{title}{Imagenet: A large-scale hierarchical image database}, in: \bibinfo{booktitle}{2009 IEEE conference on computer vision and pattern recognition}, \bibinfo{organization}{Ieee}. pp. \bibinfo{pages}{248--255}.
%Type = Article
\bibitem[{Dessardo et~al.(2019)Dessardo, Musta{\'c}, Banac and Dessardo}]{dessardo2019paths}
\bibinfo{author}{Dessardo, N.S.}, \bibinfo{author}{Musta{\'c}, E.}, \bibinfo{author}{Banac, S.}, \bibinfo{author}{Dessardo, S.}, \bibinfo{year}{2019}.
\newblock \bibinfo{title}{Paths of causal influence from prenatal inflammation and preterm gestation to childhood asthma symptoms}.
\newblock \bibinfo{journal}{J Asthma} \bibinfo{volume}{56}, \bibinfo{pages}{823--832}.
%Type = Article
\bibitem[{Ernst et~al.(2023)Ernst, Basic, Freedman, Price and Suresh}]{ernst2023comparison}
\bibinfo{author}{Ernst, L.M.}, \bibinfo{author}{Basic, E.}, \bibinfo{author}{Freedman, A.A.}, \bibinfo{author}{Price, E.}, \bibinfo{author}{Suresh, S.}, \bibinfo{year}{2023}.
\newblock \bibinfo{title}{Comparison of placental pathology reports from spontaneous preterm births finalized by general surgical pathologists versus perinatal pathologist: A call to action}.
\newblock \bibinfo{journal}{The American Journal of Surgical Pathology} \bibinfo{volume}{47}, \bibinfo{pages}{1116--1121}.
%Type = Article
\bibitem[{Ferlaino et~al.(2018)Ferlaino, Glastonbury, Motta-Mejia, Vatish, Granne, Kennedy, Lindgren and Nell{\aa}ker}]{ferlaino2018towards}
\bibinfo{author}{Ferlaino, M.}, \bibinfo{author}{Glastonbury, C.A.}, \bibinfo{author}{Motta-Mejia, C.}, \bibinfo{author}{Vatish, M.}, \bibinfo{author}{Granne, I.}, \bibinfo{author}{Kennedy, S.}, \bibinfo{author}{Lindgren, C.M.}, \bibinfo{author}{Nell{\aa}ker, C.}, \bibinfo{year}{2018}.
\newblock \bibinfo{title}{Towards deep cellular phenotyping in placental histology}.
\newblock \bibinfo{journal}{arXiv preprint arXiv:1804.03270} .
%Type = Article
\bibitem[{Filiot et~al.(2023)Filiot, Ghermi, Olivier, Jacob, Fidon, Mac~Kain, Saillard and Schiratti}]{filiot2023scaling}
\bibinfo{author}{Filiot, A.}, \bibinfo{author}{Ghermi, R.}, \bibinfo{author}{Olivier, A.}, \bibinfo{author}{Jacob, P.}, \bibinfo{author}{Fidon, L.}, \bibinfo{author}{Mac~Kain, A.}, \bibinfo{author}{Saillard, C.}, \bibinfo{author}{Schiratti, J.B.}, \bibinfo{year}{2023}.
\newblock \bibinfo{title}{Scaling self-supervised learning for histopathology with masked image modeling}.
\newblock \bibinfo{journal}{medRxiv} , \bibinfo{pages}{2023--07}.
%Type = Article
\bibitem[{Getahun et~al.(2010)Getahun, Strickland, Zeiger, Fassett, Chen, Rhoads and Jacobsen}]{getahun2010effect}
\bibinfo{author}{Getahun, D.}, \bibinfo{author}{Strickland, D.}, \bibinfo{author}{Zeiger, R.S.}, \bibinfo{author}{Fassett, M.J.}, \bibinfo{author}{Chen, W.}, \bibinfo{author}{Rhoads, G.G.}, \bibinfo{author}{Jacobsen, S.J.}, \bibinfo{year}{2010}.
\newblock \bibinfo{title}{Effect of chorioamnionitis on early childhood asthma}.
\newblock \bibinfo{journal}{Archives of pediatrics \& adolescent medicine} \bibinfo{volume}{164}, \bibinfo{pages}{187--192}.
%Type = Article
\bibitem[{Goldstein et~al.(2020)Goldstein, Gallagher, Beck, Kumar and Gernand}]{goldstein2020maternal}
\bibinfo{author}{Goldstein, J.A.}, \bibinfo{author}{Gallagher, K.}, \bibinfo{author}{Beck, C.}, \bibinfo{author}{Kumar, R.}, \bibinfo{author}{Gernand, A.D.}, \bibinfo{year}{2020}.
\newblock \bibinfo{title}{Maternal-fetal inflammation in the placenta and the developmental origins of health and disease}.
\newblock \bibinfo{journal}{Frontiers in immunology} \bibinfo{volume}{11}, \bibinfo{pages}{531543}.
%Type = Article
\bibitem[{Goldstein et~al.(2023)Goldstein, Nateghi, Irmakci and Cooper}]{goldstein2023machine}
\bibinfo{author}{Goldstein, J.A.}, \bibinfo{author}{Nateghi, R.}, \bibinfo{author}{Irmakci, I.}, \bibinfo{author}{Cooper, L.A.}, \bibinfo{year}{2023}.
\newblock \bibinfo{title}{Machine learning classification of placental villous infarction, perivillous fibrin deposition, and intervillous thrombus}.
\newblock \bibinfo{journal}{Placenta} \bibinfo{volume}{135}, \bibinfo{pages}{43--50}.
%Type = Article
\bibitem[{Irmakci et~al.(2024)Irmakci, Nateghi, Zhou, Vescovo, Saft, Ross, Yang, Cooper and Goldstein}]{irmakci2024tissue}
\bibinfo{author}{Irmakci, I.}, \bibinfo{author}{Nateghi, R.}, \bibinfo{author}{Zhou, R.}, \bibinfo{author}{Vescovo, M.}, \bibinfo{author}{Saft, M.}, \bibinfo{author}{Ross, A.E.}, \bibinfo{author}{Yang, X.J.}, \bibinfo{author}{Cooper, L.A.}, \bibinfo{author}{Goldstein, J.A.}, \bibinfo{year}{2024}.
\newblock \bibinfo{title}{Tissue contamination challenges the credibility of machine learning models in real world digital pathology}.
\newblock \bibinfo{journal}{Modern Pathology} \bibinfo{volume}{37}, \bibinfo{pages}{100422}.
%Type = Article
\bibitem[{Jung et~al.(2023)Jung, Romero, Suksai, Gotsch, Chaemsaithong, Erez, Conde-Agudelo, Gomez-Lopez, Berry, Meyyazhagan et~al.}]{jung2023clinical}
\bibinfo{author}{Jung, E.}, \bibinfo{author}{Romero, R.}, \bibinfo{author}{Suksai, M.}, \bibinfo{author}{Gotsch, F.}, \bibinfo{author}{Chaemsaithong, P.}, \bibinfo{author}{Erez, O.}, \bibinfo{author}{Conde-Agudelo, A.}, \bibinfo{author}{Gomez-Lopez, N.}, \bibinfo{author}{Berry, S.M.}, \bibinfo{author}{Meyyazhagan, A.}, et~al., \bibinfo{year}{2023}.
\newblock \bibinfo{title}{Clinical chorioamnionitis at term: definition, pathogenesis, microbiology, diagnosis, and treatment}.
\newblock \bibinfo{journal}{American journal of obstetrics and gynecology} .
%Type = Article
\bibitem[{Kim et~al.(2015)Kim, Romero, Chaemsaithong, Chaiyasit, Yoon and Kim}]{kim2015acute}
\bibinfo{author}{Kim, C.J.}, \bibinfo{author}{Romero, R.}, \bibinfo{author}{Chaemsaithong, P.}, \bibinfo{author}{Chaiyasit, N.}, \bibinfo{author}{Yoon, B.H.}, \bibinfo{author}{Kim, Y.M.}, \bibinfo{year}{2015}.
\newblock \bibinfo{title}{Acute chorioamnionitis and funisitis: definition, pathologic features, and clinical significance}.
\newblock \bibinfo{journal}{American journal of obstetrics and gynecology} \bibinfo{volume}{213}, \bibinfo{pages}{S29--S52}.
%Type = Article
\bibitem[{Kumar et~al.(2008)Kumar, Yu, Story, Pongracic, Gupta, Pearson, Ortiz, Bauchner and Wang}]{kumar2008prematurity}
\bibinfo{author}{Kumar, R.}, \bibinfo{author}{Yu, Y.}, \bibinfo{author}{Story, R.E.}, \bibinfo{author}{Pongracic, J.A.}, \bibinfo{author}{Gupta, R.}, \bibinfo{author}{Pearson, C.}, \bibinfo{author}{Ortiz, K.}, \bibinfo{author}{Bauchner, H.C.}, \bibinfo{author}{Wang, X.}, \bibinfo{year}{2008}.
\newblock \bibinfo{title}{Prematurity, chorioamnionitis, and the development of recurrent wheezing: a prospective birth cohort study}.
\newblock \bibinfo{journal}{Journal of Allergy and Clinical Immunology} \bibinfo{volume}{121}, \bibinfo{pages}{878--884}.
%Type = Article
\bibitem[{Lagodka et~al.(2022)Lagodka, Petrucci, Moretti, Cabbad and Lakhi}]{lagodka2022fetal}
\bibinfo{author}{Lagodka, S.}, \bibinfo{author}{Petrucci, S.}, \bibinfo{author}{Moretti, M.L.}, \bibinfo{author}{Cabbad, M.}, \bibinfo{author}{Lakhi, N.A.}, \bibinfo{year}{2022}.
\newblock \bibinfo{title}{Fetal and maternal inflammatory response in the setting of maternal intrapartum fever with and without clinical and histologic chorioamnionitis}.
\newblock \bibinfo{journal}{American Journal of Obstetrics \& Gynecology MFM} \bibinfo{volume}{4}, \bibinfo{pages}{100539}.
%Type = Article
\bibitem[{Laleh et~al.(2022)Laleh, Muti, Loeffler, Echle, Saldanha, Mahmood, Lu, Trautwein, Langer, Dislich et~al.}]{laleh2022benchmarking}
\bibinfo{author}{Laleh, N.G.}, \bibinfo{author}{Muti, H.S.}, \bibinfo{author}{Loeffler, C.M.L.}, \bibinfo{author}{Echle, A.}, \bibinfo{author}{Saldanha, O.L.}, \bibinfo{author}{Mahmood, F.}, \bibinfo{author}{Lu, M.Y.}, \bibinfo{author}{Trautwein, C.}, \bibinfo{author}{Langer, R.}, \bibinfo{author}{Dislich, B.}, et~al., \bibinfo{year}{2022}.
\newblock \bibinfo{title}{Benchmarking weakly-supervised deep learning pipelines for whole slide classification in computational pathology}.
\newblock \bibinfo{journal}{Medical image analysis} \bibinfo{volume}{79}, \bibinfo{pages}{102474}.
%Type = Article
\bibitem[{Lau et~al.(2005)Lau, Magee, Qiu, Houb{\'e}, Von~Dadelszen and Lee}]{lau2005chorioamnionitis}
\bibinfo{author}{Lau, J.}, \bibinfo{author}{Magee, F.}, \bibinfo{author}{Qiu, Z.}, \bibinfo{author}{Houb{\'e}, J.}, \bibinfo{author}{Von~Dadelszen, P.}, \bibinfo{author}{Lee, S.K.}, \bibinfo{year}{2005}.
\newblock \bibinfo{title}{Chorioamnionitis with a fetal inflammatory response is associated with higher neonatal mortality, morbidity, and resource use than chorioamnionitis displaying a maternal inflammatory response only}.
\newblock \bibinfo{journal}{American journal of obstetrics and gynecology} \bibinfo{volume}{193}, \bibinfo{pages}{708--713}.
%Type = Article
\bibitem[{Lu et~al.(2021)Lu, Williamson, Chen, Chen, Barbieri and Mahmood}]{lu2021data}
\bibinfo{author}{Lu, M.Y.}, \bibinfo{author}{Williamson, D.F.}, \bibinfo{author}{Chen, T.Y.}, \bibinfo{author}{Chen, R.J.}, \bibinfo{author}{Barbieri, M.}, \bibinfo{author}{Mahmood, F.}, \bibinfo{year}{2021}.
\newblock \bibinfo{title}{Data-efficient and weakly supervised computational pathology on whole-slide images}.
\newblock \bibinfo{journal}{Nature biomedical engineering} \bibinfo{volume}{5}, \bibinfo{pages}{555--570}.
%Type = Article
\bibitem[{Lynch et~al.(2018)Lynch, Berning, Thevarajah, Wagner, Post, McCourt, Cathcart, Hodges, Mandava, Gibbs et~al.}]{lynch2018role}
\bibinfo{author}{Lynch, A.M.}, \bibinfo{author}{Berning, A.A.}, \bibinfo{author}{Thevarajah, T.S.}, \bibinfo{author}{Wagner, B.D.}, \bibinfo{author}{Post, M.D.}, \bibinfo{author}{McCourt, E.A.}, \bibinfo{author}{Cathcart, J.N.}, \bibinfo{author}{Hodges, J.K.}, \bibinfo{author}{Mandava, N.}, \bibinfo{author}{Gibbs, R.S.}, et~al., \bibinfo{year}{2018}.
\newblock \bibinfo{title}{The role of the maternal and fetal inflammatory response in retinopathy of prematurity}.
\newblock \bibinfo{journal}{American Journal of Reproductive Immunology} \bibinfo{volume}{80}, \bibinfo{pages}{e12986}.
%Type = Article
\bibitem[{Maki et~al.(2022)Maki, Sato, Furukawa and Sameshima}]{maki2022histological}
\bibinfo{author}{Maki, Y.}, \bibinfo{author}{Sato, Y.}, \bibinfo{author}{Furukawa, S.}, \bibinfo{author}{Sameshima, H.}, \bibinfo{year}{2022}.
\newblock \bibinfo{title}{Histological severity of maternal and fetal inflammation is correlated with the prevalence of maternal clinical signs}.
\newblock \bibinfo{journal}{Journal of Obstetrics and Gynaecology Research} \bibinfo{volume}{48}, \bibinfo{pages}{1318--1327}.
%Type = Article
\bibitem[{Marletta et~al.(2023)Marletta, Pantanowitz, Santonicco, Caputo, Bragantini, Brunelli, Girolami and Eccher}]{marletta2023application}
\bibinfo{author}{Marletta, S.}, \bibinfo{author}{Pantanowitz, L.}, \bibinfo{author}{Santonicco, N.}, \bibinfo{author}{Caputo, A.}, \bibinfo{author}{Bragantini, E.}, \bibinfo{author}{Brunelli, M.}, \bibinfo{author}{Girolami, I.}, \bibinfo{author}{Eccher, A.}, \bibinfo{year}{2023}.
\newblock \bibinfo{title}{Application of digital imaging and artificial intelligence to pathology of the placenta}.
\newblock \bibinfo{journal}{Pediatric and Developmental Pathology} \bibinfo{volume}{26}, \bibinfo{pages}{5--12}.
%Type = Article
\bibitem[{McDowell et~al.(2016)McDowell, Jobe, Fenchel, Hardie, Gisslen, Young, Chougnet, Davis and Kallapur}]{mcdowell2016pulmonary}
\bibinfo{author}{McDowell, K.M.}, \bibinfo{author}{Jobe, A.H.}, \bibinfo{author}{Fenchel, M.}, \bibinfo{author}{Hardie, W.D.}, \bibinfo{author}{Gisslen, T.}, \bibinfo{author}{Young, L.R.}, \bibinfo{author}{Chougnet, C.A.}, \bibinfo{author}{Davis, S.D.}, \bibinfo{author}{Kallapur, S.G.}, \bibinfo{year}{2016}.
\newblock \bibinfo{title}{Pulmonary morbidity in infancy after exposure to chorioamnionitis in late preterm infants}.
\newblock \bibinfo{journal}{Annals of the American Thoracic Society} \bibinfo{volume}{13}, \bibinfo{pages}{867--876}.
%Type = Article
\bibitem[{Mobadersany et~al.(2021)Mobadersany, Cooper and Goldstein}]{mobadersany2021gestaltnet}
\bibinfo{author}{Mobadersany, P.}, \bibinfo{author}{Cooper, L.A.}, \bibinfo{author}{Goldstein, J.A.}, \bibinfo{year}{2021}.
\newblock \bibinfo{title}{Gestaltnet: aggregation and attention to improve deep learning of gestational age from placental whole-slide images}.
\newblock \bibinfo{journal}{Laboratory Investigation} \bibinfo{volume}{101}, \bibinfo{pages}{942--951}.
%Type = Article
\bibitem[{Oh et~al.(2017)Oh, Kim, Hong, Maymon, Erez, Panaitescu, Gomez-Lopez, Romero and Yoon}]{oh2017twenty}
\bibinfo{author}{Oh, K.J.}, \bibinfo{author}{Kim, S.M.}, \bibinfo{author}{Hong, J.S.}, \bibinfo{author}{Maymon, E.}, \bibinfo{author}{Erez, O.}, \bibinfo{author}{Panaitescu, B.}, \bibinfo{author}{Gomez-Lopez, N.}, \bibinfo{author}{Romero, R.}, \bibinfo{author}{Yoon, B.H.}, \bibinfo{year}{2017}.
\newblock \bibinfo{title}{Twenty-four percent of patients with clinical chorioamnionitis in preterm gestations have no evidence of either culture-proven intraamniotic infection or intraamniotic inflammation}.
\newblock \bibinfo{journal}{American journal of obstetrics and gynecology} \bibinfo{volume}{216}, \bibinfo{pages}{604--e1}.
%Type = Article
\bibitem[{Patnaik et~al.(2024)Patnaik, Khodaee, Vasam, Mukherjee, Salsabili, Ukwatta, Grynspan, Chan and Bainbridge}]{patnaik2024automated}
\bibinfo{author}{Patnaik, P.}, \bibinfo{author}{Khodaee, A.}, \bibinfo{author}{Vasam, G.}, \bibinfo{author}{Mukherjee, A.}, \bibinfo{author}{Salsabili, S.}, \bibinfo{author}{Ukwatta, E.}, \bibinfo{author}{Grynspan, D.}, \bibinfo{author}{Chan, A.D.}, \bibinfo{author}{Bainbridge, S.}, \bibinfo{year}{2024}.
\newblock \bibinfo{title}{Automated detection of microscopic placental features indicative of maternal vascular malperfusion using machine learning}.
\newblock \bibinfo{journal}{Placenta} \bibinfo{volume}{145}, \bibinfo{pages}{19--26}.
%Type = Article
\bibitem[{Rallis et~al.(2022)Rallis, Lithoxopoulou, Pervana, Karagianni, Hatziioannidis, Soubasi and Tsakalidis}]{rallis2022clinical}
\bibinfo{author}{Rallis, D.}, \bibinfo{author}{Lithoxopoulou, M.}, \bibinfo{author}{Pervana, S.}, \bibinfo{author}{Karagianni, P.}, \bibinfo{author}{Hatziioannidis, I.}, \bibinfo{author}{Soubasi, V.}, \bibinfo{author}{Tsakalidis, C.}, \bibinfo{year}{2022}.
\newblock \bibinfo{title}{Clinical chorioamnionitis and histologic placental inflammation: association with early-neonatal sepsis}.
\newblock \bibinfo{journal}{The Journal of Maternal-Fetal \& Neonatal Medicine} \bibinfo{volume}{35}, \bibinfo{pages}{8090--8096}.
%Type = Inproceedings
\bibitem[{Redline(2006)}]{redline2006inflammatory}
\bibinfo{author}{Redline, R.W.}, \bibinfo{year}{2006}.
\newblock \bibinfo{title}{Inflammatory responses in the placenta and umbilical cord}, in: \bibinfo{booktitle}{Seminars in Fetal and Neonatal Medicine}, \bibinfo{organization}{Elsevier}. pp. \bibinfo{pages}{296--301}.
%Type = Article
\bibitem[{Redline et~al.(2022)Redline, Vik, Heerema-McKenney, Jamtoy, Ravishankar, Ton~Nu, Vogt, Ng, Nelson, Lydersen et~al.}]{redline2022interobserver}
\bibinfo{author}{Redline, R.W.}, \bibinfo{author}{Vik, T.}, \bibinfo{author}{Heerema-McKenney, A.}, \bibinfo{author}{Jamtoy, A.H.}, \bibinfo{author}{Ravishankar, S.}, \bibinfo{author}{Ton~Nu, T.N.}, \bibinfo{author}{Vogt, C.}, \bibinfo{author}{Ng, P.}, \bibinfo{author}{Nelson, K.B.}, \bibinfo{author}{Lydersen, S.}, et~al., \bibinfo{year}{2022}.
\newblock \bibinfo{title}{Interobserver reliability for identifying specific patterns of placental injury as defined by the amsterdam classification}.
\newblock \bibinfo{journal}{Archives of Pathology \& Laboratory Medicine} \bibinfo{volume}{146}, \bibinfo{pages}{372--378}.
%Type = Article
\bibitem[{Redline et~al.(2000)Redline, Wilson-Costello, Borawski, Fanaroff and Hack}]{redline2000relationship}
\bibinfo{author}{Redline, R.W.}, \bibinfo{author}{Wilson-Costello, D.}, \bibinfo{author}{Borawski, E.}, \bibinfo{author}{Fanaroff, A.A.}, \bibinfo{author}{Hack, M.}, \bibinfo{year}{2000}.
\newblock \bibinfo{title}{The relationship between placental and other perinatal risk factors for neurologic impairment in very low birth weight children}.
\newblock \bibinfo{journal}{Pediatric research} \bibinfo{volume}{47}, \bibinfo{pages}{721--726}.
%Type = Article
\bibitem[{Roberts et~al.(2012)Roberts, Celi, Riley, Onderdonk, Boyd, Johnson and Lieberman}]{roberts2012acute}
\bibinfo{author}{Roberts, D.J.}, \bibinfo{author}{Celi, A.C.}, \bibinfo{author}{Riley, L.E.}, \bibinfo{author}{Onderdonk, A.B.}, \bibinfo{author}{Boyd, T.K.}, \bibinfo{author}{Johnson, L.C.}, \bibinfo{author}{Lieberman, E.}, \bibinfo{year}{2012}.
\newblock \bibinfo{title}{Acute histologic chorioamnionitis at term: nearly always noninfectious}.
\newblock \bibinfo{journal}{PloS one} \bibinfo{volume}{7}, \bibinfo{pages}{e31819}.
%Type = Article
\bibitem[{Romero et~al.(2016a)Romero, Chaemsaithong, Docheva, Korzeniewski, Kusanovic, Yoon, Kim, Chaiyasit, Ahmed, Qureshi et~al.}]{romero2016clinical2}
\bibinfo{author}{Romero, R.}, \bibinfo{author}{Chaemsaithong, P.}, \bibinfo{author}{Docheva, N.}, \bibinfo{author}{Korzeniewski, S.J.}, \bibinfo{author}{Kusanovic, J.P.}, \bibinfo{author}{Yoon, B.H.}, \bibinfo{author}{Kim, J.S.}, \bibinfo{author}{Chaiyasit, N.}, \bibinfo{author}{Ahmed, A.I.}, \bibinfo{author}{Qureshi, F.}, et~al., \bibinfo{year}{2016}a.
\newblock \bibinfo{title}{Clinical chorioamnionitis at term vi: acute chorioamnionitis and funisitis according to the presence or absence of microorganisms and inflammation in the amniotic cavity}.
\newblock \bibinfo{journal}{Journal of perinatal medicine} \bibinfo{volume}{44}, \bibinfo{pages}{33--51}.
%Type = Article
\bibitem[{Romero et~al.(2016b)Romero, Chaemsaithong, Korzeniewski, Kusanovic, Docheva, Martinez-Varea, Ahmed, Yoon, Hassan, Chaiworapongsa et~al.}]{romero2016clinical}
\bibinfo{author}{Romero, R.}, \bibinfo{author}{Chaemsaithong, P.}, \bibinfo{author}{Korzeniewski, S.J.}, \bibinfo{author}{Kusanovic, J.P.}, \bibinfo{author}{Docheva, N.}, \bibinfo{author}{Martinez-Varea, A.}, \bibinfo{author}{Ahmed, A.I.}, \bibinfo{author}{Yoon, B.H.}, \bibinfo{author}{Hassan, S.S.}, \bibinfo{author}{Chaiworapongsa, T.}, et~al., \bibinfo{year}{2016}b.
\newblock \bibinfo{title}{Clinical chorioamnionitis at term iii: how well do clinical criteria perform in the identification of proven intra-amniotic infection?}
\newblock \bibinfo{journal}{Journal of perinatal medicine} \bibinfo{volume}{44}, \bibinfo{pages}{23--32}.
%Type = Article
\bibitem[{Romero et~al.(2019)Romero, Gomez-Lopez, Winters, Jung, Shaman, Bieda, Panaitescu, Pacora, Erez, Greenberg et~al.}]{romero2019evidence}
\bibinfo{author}{Romero, R.}, \bibinfo{author}{Gomez-Lopez, N.}, \bibinfo{author}{Winters, A.D.}, \bibinfo{author}{Jung, E.}, \bibinfo{author}{Shaman, M.}, \bibinfo{author}{Bieda, J.}, \bibinfo{author}{Panaitescu, B.}, \bibinfo{author}{Pacora, P.}, \bibinfo{author}{Erez, O.}, \bibinfo{author}{Greenberg, J.M.}, et~al., \bibinfo{year}{2019}.
\newblock \bibinfo{title}{Evidence that intra-amniotic infections are often the result of an ascending invasion--a molecular microbiological study}.
\newblock \bibinfo{journal}{Journal of perinatal medicine} \bibinfo{volume}{47}, \bibinfo{pages}{915--931}.
%Type = Article
\bibitem[{Romero et~al.(2014)Romero, Miranda, Chaiworapongsa, Korzeniewski, Chaemsaithong, Gotsch, Dong, Ahmed, Yoon, Hassan et~al.}]{romero2014prevalence}
\bibinfo{author}{Romero, R.}, \bibinfo{author}{Miranda, J.}, \bibinfo{author}{Chaiworapongsa, T.}, \bibinfo{author}{Korzeniewski, S.J.}, \bibinfo{author}{Chaemsaithong, P.}, \bibinfo{author}{Gotsch, F.}, \bibinfo{author}{Dong, Z.}, \bibinfo{author}{Ahmed, A.I.}, \bibinfo{author}{Yoon, B.H.}, \bibinfo{author}{Hassan, S.S.}, et~al., \bibinfo{year}{2014}.
\newblock \bibinfo{title}{Prevalence and clinical significance of sterile intra-amniotic inflammation in patients with preterm labor and intact membranes}.
\newblock \bibinfo{journal}{American journal of reproductive immunology} \bibinfo{volume}{72}, \bibinfo{pages}{458--474}.
%Type = Article
\bibitem[{Simmonds et~al.(2004)Simmonds, Jeffery, Watson and Russell}]{simmonds2004intraobserver}
\bibinfo{author}{Simmonds, M.}, \bibinfo{author}{Jeffery, H.}, \bibinfo{author}{Watson, G.}, \bibinfo{author}{Russell, P.}, \bibinfo{year}{2004}.
\newblock \bibinfo{title}{Intraobserver and interobserver variability for the histologic diagnosis of chorioamnionitis}.
\newblock \bibinfo{journal}{American journal of obstetrics and gynecology} \bibinfo{volume}{190}, \bibinfo{pages}{152--155}.
%Type = Article
\bibitem[{Song et~al.(2023)Song, Jaume, Williamson, Lu, Vaidya, Miller and Mahmood}]{song2023artificial}
\bibinfo{author}{Song, A.H.}, \bibinfo{author}{Jaume, G.}, \bibinfo{author}{Williamson, D.F.}, \bibinfo{author}{Lu, M.Y.}, \bibinfo{author}{Vaidya, A.}, \bibinfo{author}{Miller, T.R.}, \bibinfo{author}{Mahmood, F.}, \bibinfo{year}{2023}.
\newblock \bibinfo{title}{Artificial intelligence for digital and computational pathology}.
\newblock \bibinfo{journal}{Nature Reviews Bioengineering} \bibinfo{volume}{1}, \bibinfo{pages}{930--949}.
%Type = Article
\bibitem[{Srinidhi et~al.(2021)Srinidhi, Ciga and Martel}]{srinidhi2021deep}
\bibinfo{author}{Srinidhi, C.L.}, \bibinfo{author}{Ciga, O.}, \bibinfo{author}{Martel, A.L.}, \bibinfo{year}{2021}.
\newblock \bibinfo{title}{Deep neural network models for computational histopathology: A survey}.
\newblock \bibinfo{journal}{Medical image analysis} \bibinfo{volume}{67}, \bibinfo{pages}{101813}.
%Type = Article
\bibitem[{Straughen et~al.(2017)Straughen, Misra, Divine, Shah, Perez, VanHorn, Onbreyt, Dygulska, Schmitt, Lederman et~al.}]{straughen2017association}
\bibinfo{author}{Straughen, J.K.}, \bibinfo{author}{Misra, D.P.}, \bibinfo{author}{Divine, G.}, \bibinfo{author}{Shah, R.}, \bibinfo{author}{Perez, G.}, \bibinfo{author}{VanHorn, S.}, \bibinfo{author}{Onbreyt, V.}, \bibinfo{author}{Dygulska, B.}, \bibinfo{author}{Schmitt, R.}, \bibinfo{author}{Lederman, S.}, et~al., \bibinfo{year}{2017}.
\newblock \bibinfo{title}{The association between placental histopathology and autism spectrum disorder}.
\newblock \bibinfo{journal}{Placenta} \bibinfo{volume}{57}, \bibinfo{pages}{183--188}.
%Type = Inproceedings
\bibitem[{Tan and Le(2021)}]{tan2021efficientnetv2}
\bibinfo{author}{Tan, M.}, \bibinfo{author}{Le, Q.}, \bibinfo{year}{2021}.
\newblock \bibinfo{title}{Efficientnetv2: Smaller models and faster training}, in: \bibinfo{booktitle}{International conference on machine learning}, \bibinfo{organization}{PMLR}. pp. \bibinfo{pages}{10096--10106}.
%Type = Book
\bibitem[{Van~Rossum(2020)}]{van1995python}
\bibinfo{author}{Van~Rossum, G.}, \bibinfo{year}{2020}.
\newblock \bibinfo{title}{The Python Library Reference, release 3.8.2}.
\newblock \bibinfo{publisher}{Python Software Foundation}.
%Type = Article
\bibitem[{Vanea et~al.(2022)Vanea, D{\v{z}}igurski, Rukins, Dodi, Siigur, Salum{\"a}e, Meir, Parks, Hochner-Celnikier, Fraser et~al.}]{vanea2022happy}
\bibinfo{author}{Vanea, C.}, \bibinfo{author}{D{\v{z}}igurski, J.}, \bibinfo{author}{Rukins, V.}, \bibinfo{author}{Dodi, O.}, \bibinfo{author}{Siigur, S.}, \bibinfo{author}{Salum{\"a}e, L.}, \bibinfo{author}{Meir, K.}, \bibinfo{author}{Parks, W.}, \bibinfo{author}{Hochner-Celnikier, D.}, \bibinfo{author}{Fraser, A.}, et~al., \bibinfo{year}{2022}.
\newblock \bibinfo{title}{Happy: A deep learning pipeline for mapping cell-to-tissue graphs across placenta histology whole slide images} .
%Type = Article
\bibitem[{Xiao et~al.(2018)Xiao, Zhu, Qu, Gou, Huang, Li and Mu}]{xiao2018maternal}
\bibinfo{author}{Xiao, D.}, \bibinfo{author}{Zhu, T.}, \bibinfo{author}{Qu, Y.}, \bibinfo{author}{Gou, X.}, \bibinfo{author}{Huang, Q.}, \bibinfo{author}{Li, X.}, \bibinfo{author}{Mu, D.}, \bibinfo{year}{2018}.
\newblock \bibinfo{title}{Maternal chorioamnionitis and neurodevelopmental outcomes in preterm and very preterm neonates: a meta-analysis}.
\newblock \bibinfo{journal}{PLoS One} \bibinfo{volume}{13}, \bibinfo{pages}{e0208302}.
%Type = Article
\bibitem[{Zaidi et~al.(2020)Zaidi, Lamalmi, Lahlou, Slaoui, Barkat, Alamrani and Alhamany}]{zaidi2020clinical}
\bibinfo{author}{Zaidi, H.}, \bibinfo{author}{Lamalmi, N.}, \bibinfo{author}{Lahlou, L.}, \bibinfo{author}{Slaoui, M.}, \bibinfo{author}{Barkat, A.}, \bibinfo{author}{Alamrani, S.}, \bibinfo{author}{Alhamany, Z.}, \bibinfo{year}{2020}.
\newblock \bibinfo{title}{Clinical predictive factors of histological chorioamnionitis: case-control study}.
\newblock \bibinfo{journal}{Heliyon} \bibinfo{volume}{6}.
%Type = Article
\bibitem[{Zhang et~al.(2020)Zhang, Davaasuren, Wu, Goldstein, Gernand and Wang}]{zhang2020multi}
\bibinfo{author}{Zhang, Z.}, \bibinfo{author}{Davaasuren, D.}, \bibinfo{author}{Wu, C.}, \bibinfo{author}{Goldstein, J.A.}, \bibinfo{author}{Gernand, A.D.}, \bibinfo{author}{Wang, J.Z.}, \bibinfo{year}{2020}.
\newblock \bibinfo{title}{Multi-region saliency-aware learning for cross-domain placenta image segmentation}.
\newblock \bibinfo{journal}{Pattern recognition letters} \bibinfo{volume}{140}, \bibinfo{pages}{165--171}.
%Type = Article
\bibitem[{Zimmermann et~al.(2024)Zimmermann, Vorontsov, Viret, Casson, Zelechowski, Shaikovski, Tenenholtz, Hall, Fuchs, Fusi et~al.}]{zimmermann2024virchow}
\bibinfo{author}{Zimmermann, E.}, \bibinfo{author}{Vorontsov, E.}, \bibinfo{author}{Viret, J.}, \bibinfo{author}{Casson, A.}, \bibinfo{author}{Zelechowski, M.}, \bibinfo{author}{Shaikovski, G.}, \bibinfo{author}{Tenenholtz, N.}, \bibinfo{author}{Hall, J.}, \bibinfo{author}{Fuchs, T.}, \bibinfo{author}{Fusi, N.}, et~al., \bibinfo{year}{2024}.
\newblock \bibinfo{title}{Virchow 2: Scaling self-supervised mixed magnification models in pathology}.
\newblock \bibinfo{journal}{arXiv preprint arXiv:2408.00738} .

\end{thebibliography}


% Author biography without author photo.
% Author biography. Author biography. Author biography.

% \bio{}%figs/cas-pic1}
% Author biography with author photo.
% Author biography. Author biography. Author biography.

% \bio{}%figs/cas-pic1}
% Author biography with author photo.
% Author biography. Author biography. Author biography.

\end{document}